\documentclass[conference,compsoc]{IEEEtran}
\ifCLASSOPTIONcompsoc
  \usepackage[nocompress]{cite}
\else
  \usepackage{cite}
\fi
\usepackage{graphicx}
\usepackage{float}
\usepackage{listings}
\usepackage{subcaption}
\usepackage{multirow}
\usepackage{tikz}
\usetikzlibrary{decorations.pathreplacing,angles,quotes}
\usepackage{amsmath}
\DeclareMathOperator*{\argmax}{\arg\!\max}

\usepackage{algorithm, algpseudocode}
\usepackage{array}
\usepackage{tabu}
\usepackage{tabularx,booktabs} 
\usepackage{adjustbox}

\makeatletter
\renewcommand{\@IEEEsectpunct}{\ }
\makeatother
\DeclareMathOperator{\sign}{sgn}
\usepackage{hyperref}

\usepackage{titlesec}
\titleformat*{\paragraph}{\bfseries}

\titleformat{\paragraph}[runin]
        {\normalfont\bfseries}
        {\theparagraph.}
        {0.5em}
        {}
        [\;]
\titleformat{\subsubsection}[runin]
        {\normalfont\bfseries}
        {\thesubsubsection.}
        {0.5em}
        {}
        [\;]
\titlespacing{\paragraph}{0.0em}{.5ex}{.5ex} 

\ifCLASSINFOpdf
\else
\fi
\begin{document}
%
\title{Classification of Phonological Parameters in Sign Languages}

\author{\IEEEauthorblockN{Boris Mocialov}
\IEEEauthorblockA{School of Engineering and\\Physical Sciences\\
Heriot-Watt University\\
Edinburgh, UK\\
Email: bm4@hw.ac.uk}
\and
\IEEEauthorblockN{Graham Turner}
\IEEEauthorblockA{School of Social Sciences\\
Heriot-Watt University\\
Edinburgh, UK\\
Email: g.h.turner@hw.ac.uk}
\and
\IEEEauthorblockN{Helen Hastie}
\IEEEauthorblockA{School of Mathematics and\\Computer Science\\
Heriot-Watt University\\
Edinburgh, UK\\
Email: h.hastie@hw.ac.uk}}


%


\maketitle

\begin{abstract}
Signers compose sign language phonemes that enable communication by combining phonological parameters such as handshape, orientation, location, movement, and non-manual features. Linguistic research often breaks down signs into their constituent parts to study sign languages and often a lot of effort is invested into the annotation of the videos. In this work we show how a single model can be used to recognise the individual phonological parameters within sign languages with the aim of either to assist linguistic annotations or to describe the signs for the sign recognition models. We use Danish Sign Language data set `Ordbog over Dansk Tegnsprog' to generate multiple data sets using pose estimation model, which are then used for training the multi-label Fast R-CNN model to support multi-label modelling. Moreover, we show that there is a significant co-dependence between the orientation and location phonological parameters in the generated data and we incorporate this co-dependence in the model to achieve better performance. 
\end{abstract}


%
\IEEEpeerreviewmaketitle

\section{Introduction}
Sign languages worldwide can be described using a fixed set of phonological parameters, such as a shape of the hand, extended finger orientation, relative hand location to the body, hand movement type, and non-manual features. Some research also takes into account hand arrangement, which considers location of the hands relative to one another \cite{stokoe2005sign,sutton-spence_woll_1999}.

This work focuses on the data-driven modelling of phonological parameters (orientation, location, and handshape) from the raw single-camera images, leaving movement parameter modelling for the future work. We hypothesise that by explicitly introducing co-dependence among the phonological parameters, we could improve the performance of the overall model as is suggested by Awad et al. \cite{5414159}.

We use Danish Sign Language data set `Ordbog over Dansk Tegnsprog' \cite{troelsgaard2008electronic} to generate multiple data sets using pose estimation model, which are then used for training the Fast R-CNN model \cite{girshick2013rich} that was modified for this work.

This research is aiming to assist linguistic studies on sign languages by providing an automated annotation tool that could be used for extracting phonological parameters from the raw videos for any sign language. In addition, correct classification of the phonological parameters could also render sign recognition models more accurate if the models rely on modelling the underlying phonemes.

\subsection{Hamburg Sign Language Notation System}\label{sec:hamnosys}
Hamburg Sign Language Notation System (HamNoSys) \cite{hanke2004hamnosys} is a writing system for the phonological parameters that can be applied to any sign language. In this research, HamNoSys provides a framework for categorising categories of each phonological parameter.

\begin{table}[H]
\centering
\caption{Approximate amount of notations for every phonological parameter in HamNoSys notation system. The amount is approximate because the notation system defines combinations of potential configurations or actions for each parameter}
\resizebox{0.8\columnwidth}{!}{%
\begin{tabular}{ l| l| c }
 \multirow{2}{*}{\textbf{\shortstack[c]{Phonological\\Parameter}}} & \multirow{2}{*}{\textbf{\shortstack[c]{HamNoSys Phonological\\Parameter Sub-type}}} & \multirow{2}{*}{\shortstack[c]{\textbf{Approximate}\\\textbf{amount}}} \\ 
 & & \\ \hline
 Handshape & Hand shapes & 72 \\ \hline
 \multirow{ 2}{*}{Orientation} & Extended finger directions & 18 \\
 & Palm orientations & 8 \\ \hline
 \multirow{ 3}{*}{Location} & Hand locations & 46 \\
 & Hand location sides & 5 \\
 & Hand distances & 5 \\ \hline
 \multirow{ 5}{*}{Movement} & Hand movements & 7 \\
 & Other movements & 1 \\
 & Movement directions & 6 \\
 & Movement speeds & 5 \\
 & Movement repetitions & 7 \\ \hline
 & \multirow{3}{*}{\shortstack[l]{Eye gaze\\Facial expression\\Mouth gestures}} &  \\ 
 Non-manuals & & 12 \\
 & & \\ \hline
 \multicolumn{3}{c}{..and many other special cases}
\end{tabular}
}
\label{tab:hamnosys}
\end{table}
Table \ref{tab:hamnosys} shows the approximate number for every HamNoSys sub-type. Annotation of the non-manuals is still limited as it potentially is much more complicated and subtle than other parameters. Despite the fact that the non-manual features, such as facial gestures, play an essential part in interpretation of the sign languages, HamNoSys has relatively poor notation system for them. Therefore, non-manuals will be left out for much later future work.

\begin{figure}[H]
\centering
    \begin{tikzpicture}
    \node[anchor=south west,inner sep=0] at (0,1) {\includegraphics[width=0.2\textwidth,trim={0cm 0cm 0cm 0cm},clip]{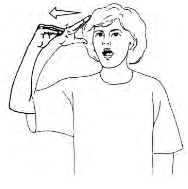}};
    
    \node[anchor=south west,inner sep=0] at (0,0) {\includegraphics[width=0.22\textwidth,trim={0cm 6.45cm 0cm 0cm},clip]{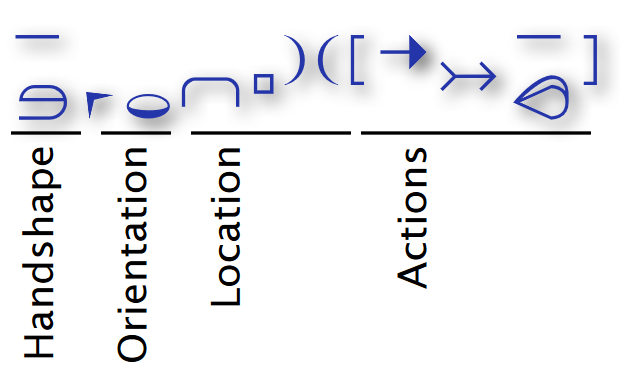}};
\draw[decoration={brace,mirror,raise=5pt},decorate]  (0,0) -- node[below=6pt] {\rotatebox[origin=c]{90}{\scriptsize Handshape}} (0.6,0);
\draw[decoration={brace,mirror,raise=5pt},decorate]  (0.6,0) -- node[below=6pt] {\rotatebox[origin=c]{90}{\scriptsize Orientation}} (1.2,0);
\draw[decoration={brace,mirror,raise=5pt},decorate]  (1.2,0) -- node[below=6pt] {\rotatebox[origin=c]{90}{\scriptsize Location}} (2.3,0);
\draw[decoration={brace,mirror,raise=5pt},decorate]  (2.3,0) -- node[below=6pt] {\rotatebox[origin=c]{90}{\scriptsize Movement}} (3.8,0);
\end{tikzpicture}
\caption{Example of a sign (`\textsc{Hamburg}') in German Sign Language and its HamNoSys notation (Source: HamNoSys documentation)}
\label{fig:sl_example}
\end{figure}
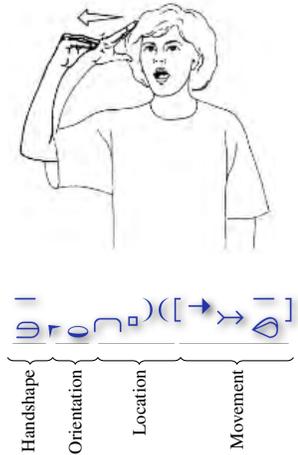

Figure \ref{fig:sl_example} shows an example of a sign `Hamburg' in German Sign Language (GSL), described using the HamNoSys notation. The notation system does not require the annotator to use all the HamNoSys sub-types and the annotation is usually task-specific \cite{hanke2004hamnosys}.


\subsection{Report Structure}
First, this paper presents the past research on the individual phonological parameters modelling. Second it describes what data was used and the generated data sets. Third, approaches to individual phonological parameter recognition are presented, while different recognition methods for the handshape parameter are compared. Fourth, the existent co-dependence between the individual parameters is calculated. Fifth, the paper presents a single approach for modelling multiple phonological parameters wile sharing learned visual features and explicitly influencing each other as well as compares that method to modelling multiple parameters that were trained separately. Finally, a model with the most optimal configuration is trained to achieve the best results. the extension of this work is presented.

%
%
%

\section{Related Work}\label{sec:related_work}

Research in automated sign language understanding focuses on two areas, applying computer vision techniques for pose estimation and tracking \cite{7850183,Mocialov2017TowardsCS} and natural language understanding techniques for sign language modelling at the gloss level (interpretation of a sign in written language) either using n-grams \cite{DBLP:journals/corr/CateH17} or neural networks more recently \cite{mocialov2018transfer}. 

Recognition \cite{7850183} and tracking of the body parts for the sign languages raise a noteworthy occlusion problem to the algorithms that are being used \cite{10.1007/978-3-642-35749-7_18}. The problem can be approached from different sides. On one hand, data-driven approaches that predict the anatomic features are trained on both raw image data together with the some sensory data which is more resilient to occlusions, such as motion capture devices \cite{jantunen2012experiences} or RGB-D sensors \cite{Mocialov2017TowardsCS}. On the other hand, approaches that focus on tracking the occluded regions of interest improve the recognition \cite{1698877}, but not by a lot \cite{cooper2007large}. In our work, we use the OpenPose library \cite{cao2018openpose,simon2017hand,cao2017realtime,wei2016cpm} for generating the datasets for our multi-label model. The library detects 2D or 3D anatomical key-points, associated with the human body, hands, face, and feet on a single image. The library provides $21$ (x,y) key-points for every part of the hand, $25$ key-points for the whole body skeleton, and $70$ key-points for the face. The library also gives a confidence value for every (x,y) pair, but this was not considered during this work. It can also provide person tracking and can be integrated with the Unity game engine.

Multiple previous works have modelled individual phonological parameters taking Stokoe's visual taxonomy as the basis \cite{stokoe2005sign}. Similar idea of modelling individual phonological parameters and constructing linguistic feature vectors has been used for recognising individual signs in \cite{10.1007/978-3-540-24670-1_30}. Their work operates on handshape, location, and movement by modelling them as Markov chains using a single example. They only provide accuracy for the handshape classifier to be $75\%$ for eight handshapes from the British Sign Language (BSL). They use sliding window to get the highest activation throughout the temporal dimension as a way to spot and classify individual signs during continuous signing. Cooper and Bowden \cite{cooper2007large} modelled location, movement, and hand arrangement and called them the sub-sign units. They have showed that tracking of the phonological parameters does not contribute much to the sign classification accuracy and provide location accuracy to be $31\%$ using the AdaBoost classifier after they apply the grid on the image and see which part of the grid fires when a hand is close to some body part. Cooper et al. \cite{cooper2012sign} relied on handshape, location, movement, and hand arrangement in their work on recognition of the individual signs in the BSL using the random forest model trained on the histogram of oriented gradients (HOG). They report confusion matrix for handshape without reporting the overall accuracy. The confusion table shows quite high recognition accuracy of the three out of twelve hand shapes and quite poor performance for the other three hand shapes. They also showed that location information contributes the most to the recognition of a sign while handshape has the least effect. Buehler et al. \cite{buehler2009learning} resorted to movement, handshape, and orientation while matching the phonemes to find similar signs. Buehler et al. \cite{buehler2010employing} used location and handshape in the multiple instance learning problem. Koller et al. \cite{7780781} focuses on three data sets (Danish, New Zealand, and German Sign Languages) and sixty hand shapes. The model is a chain of the convolutional neural networks (VGG) pre-trained on the ImageNet data \cite{simonyan2014deep}. After fine-tuning the pre-trained model on one million cropped images of sixty different hand shapes, the model achieves $63\%$ accuracy.

\bgroup
\def\arraystretch{1.5}
\begin{table}[H]
\centering
\caption{Reported results (in \%) for phonological parameters modelling in related work}
\begin{tabular}{ |l|c|c|}
\hline
 \multirow{2}{*}{\shortstack[c]{\textbf{Phonological}\\\textbf{Parameter}}} & \multirow{2}{*}{\shortstack[c]{\textbf{Result (in \%)}}} & \multirow{2}{*}{\shortstack[c]{\textbf{Data set}}} \\
 & & \\\hline
 Handedness & \multicolumn{2}{c|}{---} \\ \hline
 \multirow{2}{*}{\shortstack[c]{Handshape}} & \multirow{2}{*}{\shortstack[c]{$75$\qquad\qquad (Bowden et al. \cite{10.1007/978-3-540-24670-1_30}),\\$63$\qquad\qquad\; (Koller et al. \cite{7780781})}} &  \\
 & & \\ \hline
 Orientation & \multicolumn{2}{c|}{---} \\ \hline
 Location & $31$\;\; (Cooper and Bowden \cite{cooper2007large}) & \\ \hline
\end{tabular}
\label{tbl:related_work}
\end{table}
\egroup

From the past research, it can be seen that different phonological parameters were used to represent signs. From the Table \ref{tbl:related_work} we see that most research does not report recognition accuracy of the individual phonological parameters, rather jump directly into classifying or clustering the signs. Also, very few have reported the accuracies of their models with phonological parameters such as orientation and handedness not being modelled at all. As individual phonological parameters are executed in parallel during signing, there is a chance that they are also co-dependent. This means that when one parameter is in a certain range (e.g. hand is northbound) the co-dependent parameter can only have a limited range of possible configurations (e.g. hand is around the upper body) and vise versa. Research in sign language modelling ignores the potential this co-dependence has on the accuracy of the models. This work attempts to exploit this co-dependence for the model improvement. To the best of our knowledge, there is very little research that pays attention to relationships between phonological parameters. Awad et al. \cite{5414159} comes close to what we try to achieve by sharing features among different phonological parameters, but they do not tell explicitly which parameters could benefit from the shared features.

\section{Data}
Danish Sign Language data set `Ordbog over Dansk Tegnsprog' (OODT) data set is a digital dictionary with a web interface that allows searching for a specific sign using phonological parameters and a gloss in Danish written language. 

{\fontsize{7}{4}\selectfont
\begin{lstlisting}[caption={One Sample from the OODT data set, containing one set ($SeqNo=1$) of phonological parameters},label=tegnsprag_listing,captionpos=b]
<Entry>
  <EntryNo>7</EntryNo>
  <Gloss>TAPPE-VIDEO</Gloss>
  <SignVideo>t_2542.mp4</SignVideo>
  <Phonology>
    <Seq>
      <SeqNo>1</SeqNo>
      <SignType>2-hand parallel</SignType>
      <Handshape1>paedagog-hand aben</Handshape1>
      <HandshapeFinal>paedagog-hand</HandshapeFinal>
      <OrientationFingers>skrat frem op</OrientationFingers>
      <OrientationPalm>op</OrientationPalm>
      <Location>neutralt rum</Location>
      <Movement>ned</Movement>
      <Relation>ved siden af</Relation>
      <Repeat/>
    </Seq>
  </Phonology>
</Entry>
\end{lstlisting}
}

Listing \ref{tegnsprag_listing} shows a single entry from the annotation file that contains all the information about the data set. Each entry contains such information as the gloss, path to a video clip, handshape, orientation, location, and movement. It is important to note that the annotation does not have the exact timing information about when a certain phoneme is being used in a clip. Since some signs require multiple phonemes (just like the words can have multiple phonemes in spoken languages), every phoneme can have multiple sequences with different phonological parameters in every sequence. We were interested in the entries, which have a single sequence with one set of phonological parameters since we do not have a mechanism for segmenting changes in phonological parameters during the execution of a sign.

The OODT data set contains isolated videos of people signing one sign at a time without any additional content or context. Therefore, every video begins with signers being in resting position (having their hands down at the abdominal level or outside the frame) and end with the same position. Since the annotation does not provide any information about the exact timing when a certain phoneme is taking place in the video, we have to filter the generated data set to exclude the frames that are recorded while the signers are in the resting position.

\begin{figure}[H]
\centering
    \begin{tikzpicture}
    \node[anchor=south west,inner sep=0] at (0,0) {\includegraphics[width=0.4\textwidth,trim={0cm 3cm 0cm 0cm},clip]{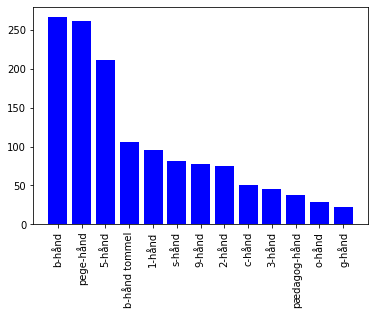}};
    
    \node[anchor=south west,inner sep=0] at (0,-2.3) {\includegraphics[width=0.4\textwidth,trim={0cm 0cm 0cm 8.1cm},clip]{pics/dataset_videos}};

    
    \node[anchor=south west,inner sep=0] at (0.95,-0.7) {\includegraphics[width=0.33\textwidth,trim={0cm 8cm 0cm 7cm},clip]{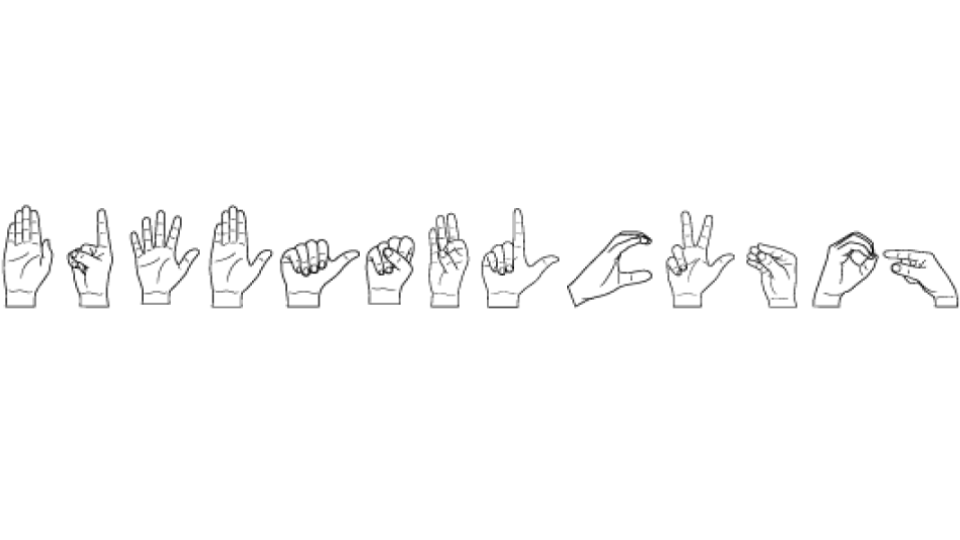}};
    
    \node (example-align) [rotate=90, align=left, text width=1.5cm] at (1.1,-3) {\scriptsize{flat hand}};

    \node (example-align) [rotate=90, align=left, text width=1.5cm] at (1.55,-3) {\scriptsize{one finger}};
    
    \node (example-align) [rotate=90, align=left, text width=1.5cm] at (2,-3) {\scriptsize{3-5 fingers}};
    
    \node (example-align) [rotate=90, align=left, text width=1.5cm] at (2.45,-3) {\scriptsize{flat hand}};
    
    \node (example-align) [rotate=90, align=left, text width=1.5cm] at (2.9,-3) {\scriptsize{tied hand}};

    \node (example-align) [rotate=90, align=left, text width=1.5cm] at (3.35,-3) {\scriptsize{tied hand}};
    
    \node (example-align) [rotate=90, align=left, text width=1.5cm] at (3.8,-3) {\scriptsize{closed hand}};
    
    \node (example-align) [rotate=90, align=left, text width=1.5cm] at (4.25,-3) {\scriptsize{two fingers}};
    
    \node (example-align) [rotate=90, align=left, text width=1.5cm] at (4.7,-3) {\scriptsize{flat hand}};
    
    \node (example-align) [rotate=90, align=left, text width=1.5cm] at (5.15,-3) {\scriptsize{3-5 fingers}};
    
    \node (example-align) [rotate=90, align=left, text width=1.5cm] at (5.6,-3) {\scriptsize{flat hand}};
    
    \node (example-align) [rotate=90, align=left, text width=1.5cm] at (6.05,-3) {\scriptsize{closed hand}};
    
    \node (example-align) [rotate=90, align=left, text width=1.5cm] at (6.5,-3) {\scriptsize{two figners}};
    
    \node [very thick] at (-0.2,1.7) {\rotatebox[origin=c]{90}{\shortstack[c]{\scriptsize Number of videos}}};
    
    \node[very thick] at (0,-3.3) {\rotatebox[origin=c]{90}{\shortstack[c]{\scriptsize Handshape\\\scriptsize Group:}}};

\end{tikzpicture}
\caption{Number of selected videos with one phoneme for each handshape for the whole OODT data set}
\label{fig:video_stats}
\end{figure}

Figure \ref{fig:video_stats} shows the handshapes, which were selected based on the number of videos in the overall data set as well as the coverage of all the hand shape groups in the OODT, which are the tied hand (s-hand, 1-hand), flat hand (b-hand, b-hand tommel, c-hand, paedagog-hand), 1-finger (pege-hand), 2-fingers (2-hand, g-hand), 3-5 fingers (3-hand, 5-hand), and closed-hand (9-hand, o-hand). 

\subsection{Pre-processing}
Pre-processing of the OODT data is done to extract frames from the raw videos, identify pose and hand information in each frame and to eliminate the frames that do not correspond to any phonemes.

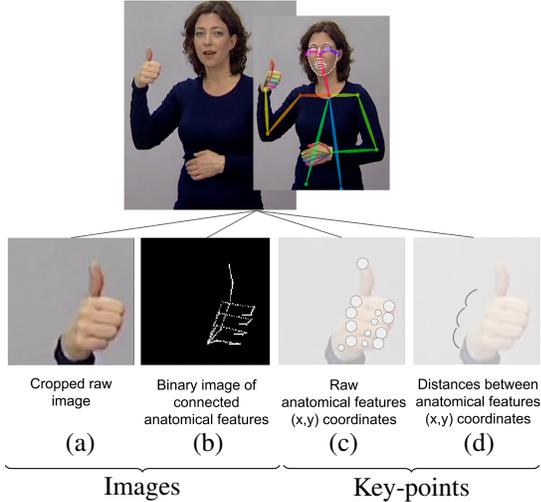
\begin{figure}[H]
\centering
    \begin{tikzpicture}
    \node[anchor=south west,inner sep=0] at (0,0) {\includegraphics[width=0.4\textwidth]{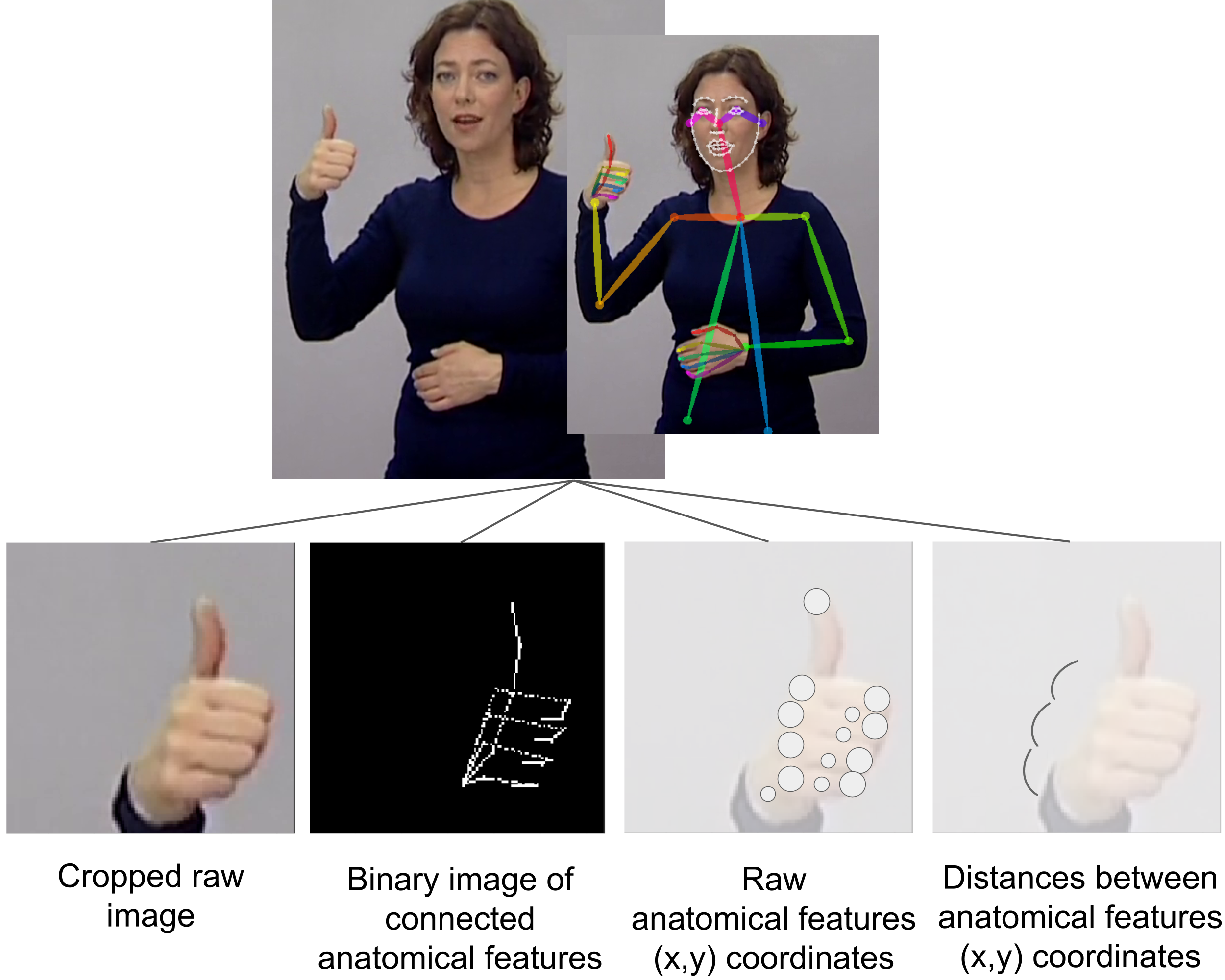}};
\draw[decoration={brace,mirror,raise=5pt},decorate]  (0,-0.3) -- node[below=6pt] {Images} (3.65,-0.3);
\draw[decoration={brace,mirror,raise=5pt},decorate]  (3.7,-0.3) -- node[below=6pt] {Key-points} (7.1,-0.3);

\node[very thick] at (1,-0.2) {(a)};
\node[very thick] at (2.7,-0.2) {(b)};
\node[very thick] at (4.5,-0.2) {(c)};
\node[very thick] at (6.3,-0.2) {(d)};

\end{tikzpicture}
\caption{Generated data sets from the original OODT data grouped into images and key-points groups. Data sets from the images group corresponding to images while the key-points group contains either raw key-points, provided by the OpenPose library or the distances between the raw key-points}
\label{fig:generated_ds}
\end{figure}

Figure \ref{fig:generated_ds} shows the generated data sets from the original OODT data. On the top is a single frame from the original data and to its right is the output from the OpenPose library. Using this information, four data sets were generated to test which features lead to a more accurate model for handshape recognition. 

First data set consists of cropped ($128\times128$ pixels) raw images of each hand as seen in Figure 8a.

Second data set consists of binary image of connected anatomic features of each hand as seen in Figure 8b. The anatomic features, provided by the OpenPose library, are connected using linear regression
\begin{gather*}
p(x) = a_0 + a_1x \\
S_r = \sum_{i=0}^{n} \lvert p(x_j) - y_j \rvert^2
 \end{gather*}
where we try to find $a_0$ (where the line intersects the axis) and $a_1$ (the slope of the line) while minimising the sum of the square of the residuals $S_r$ from the two data points. Later, we fill the blanks between the two points for steps $k$, which is chosen to be $10$.

\begin{table}[H]
\centering
\begin{tabular}{ |l| c c c c c| }
\hline
 \textbf{Type} & & & \textbf{Hand Part $(x,y)$} & & \\ \hline
 phalanx & thumb & index & middle & ring & little\\
 proximal & thumb & index & middle & ring & little \\
 metacarpals & thumb & index & middle & ring & little\\
 carpals & & index & middle & ring & little\\
 other & radius & trapezium & & &  \\ \hline
\end{tabular}
\caption{Third data set consists of the 21 (x,y) coordinates of the anatomic features provided by the OpenPose library}
\label{tbl:raw_features}
\end{table}
Third data set consists of raw (x,y) coordinates of the anatomic features as seen in Figure 8c and as described in Table \ref{tbl:raw_features}.

\begin{table}[H]
\centering
\begin{tabular}{ |l| c c c c c| }
\hline
 \textbf{From ($x_1$, $y_1$)} & & & \textbf{To ($x_2$, $y_2$)} & & \\ \hline
 radius & thumb & index & middle &  ring & little \\
 thumb &  & index & middle &  ring & little \\
 index &  &  & middle & ring & little\\
 middle &  &  &  & ring & little\\
 ring & &  &  &  & little \\\hline
\end{tabular}
\caption{Fourth data set consists of 15 distances between $(x_1,y_1)$ and $(x_2,y_2)$ coordinates of the anatomic features provided by the OpenPose library}
\label{tbl:distance_features}
\end{table}
Fourth data set consists of the 15 distances (in pixels) between the raw coordinates of the anatomic features ($x_1$,$y_1$) and ($x_2$,$y_2$) as seen in Figure 8d and as described in Table \ref{tbl:distance_features}.

To discard the frames where the signers have their hands in the resting position, we have used the hypothesis that the hand movement speed differs between the phonemes and the epentheses (hand movements between signs) \cite{choudhury2017movement,Yasugahira:2009:AHM:1667780.1667849}. We have used a sliding window over a set of 3 frames and calculated the number of pixels that the centroid of the hand has moved during the frames inside the sliding window, where the centroids are the averages of all the points $N$ provided by the OpenPose library
\begin{gather*}
centroid_{right} = (\sum_{i=1}^{N} x_{i_{right}} / N, \sum_{}^{} y_{i_{right}} / N) \\
centroid_{left} = (\sum_{i=1}^{N} x_{i_{left}} / N, \sum_{}^{} y_{i_{left}} / N) 
\end{gather*}

Later, a rectangular bounding box over a polygon that inscribes the trajectory created by the centroids of each hand over the 3 frames was generated using a third party library\footnote{\scriptsize{\url{https://bitbucket.org/william_rusnack/minimumboundingbox/src/master/}}} and the largest side of the bounding box was taken to describe the distance travelled by each hand during a window.

\begin{figure}[H]
\centering
    \begin{tikzpicture}
    \node[anchor=south west,inner sep=0] at (0,0) {\includegraphics[width=0.9\linewidth,trim={28cm 44cm 22cm 19.5cm},clip]{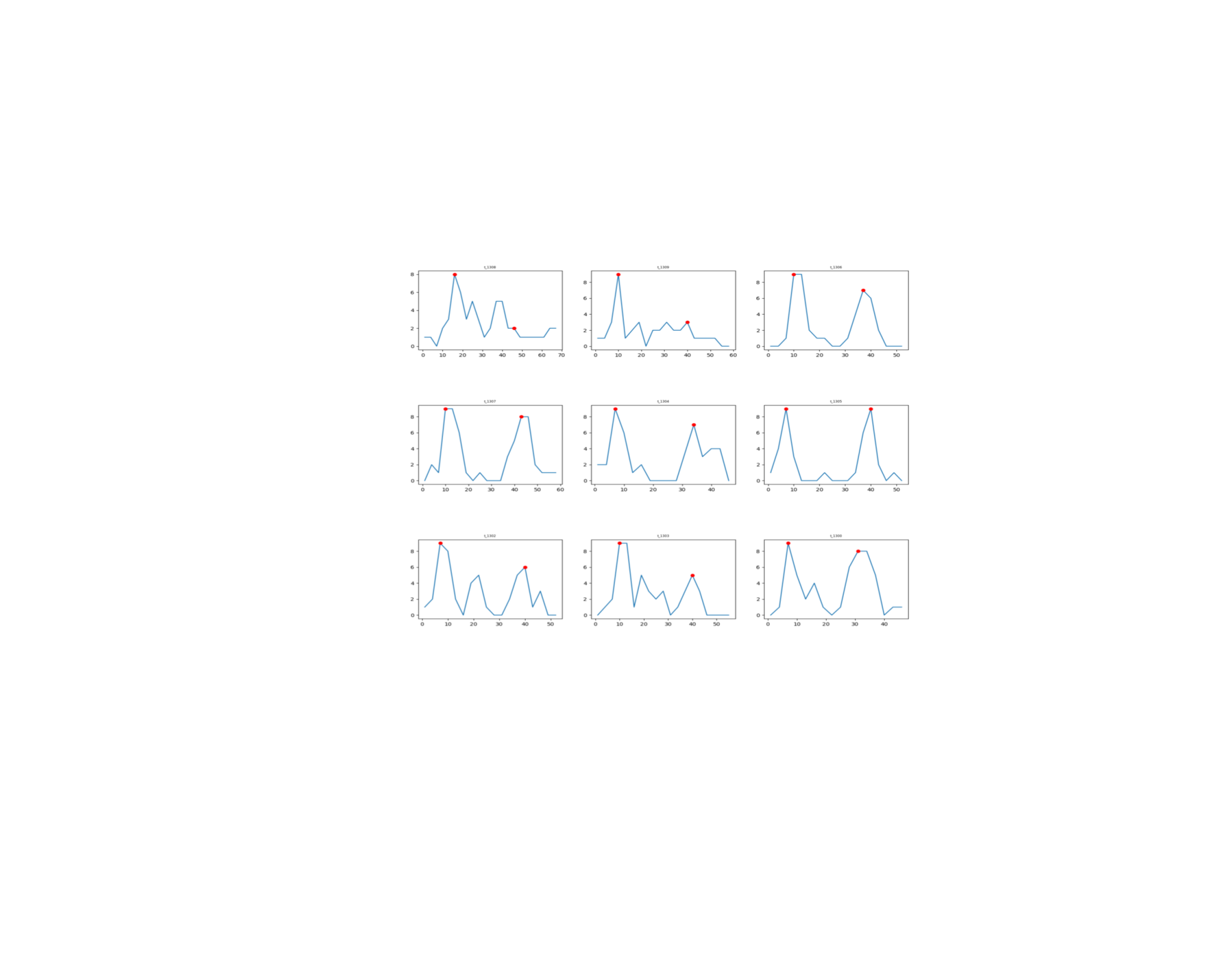}};

    
    
    \draw[red] (0.7,1.1) -- (1.2,1.6);
    \draw[green] (0.7,1.6) -- (1.2,1.1);
    
    \node[very thick] at (-0.2,0.8) {\rotatebox[origin=c]{90}{\shortstack[c]{\scriptsize Speed\\ \scriptsize (in pixels)}}};
    \node[very thick] at (4,-0.2) {\scriptsize Frame Number};
    
    \node[very thick] at (1.6,1.7) {\scriptsize Video N};
    \node[very thick] at (4,1.7) {\scriptsize Video N+1};
    \node[very thick] at (6.4,1.7) {\scriptsize Video N+2};
    
    \node[very thick] at (0.6,1.5) {\tiny $m_1$};
    \node[very thick] at (1,1.5) {\tiny $m_2$};
    
\end{tikzpicture}
\caption{Right hand centroid speeds (in pixels) for 3 arbitrary videos ($N\dots N+2$). Red dots indicate the potential start and the end of an actual sign. $m_1$ and $m_2$ indicate the slopes of the curve at consecutive frames $t_1$ and $t_2$}
\label{fig:movement}
\end{figure}
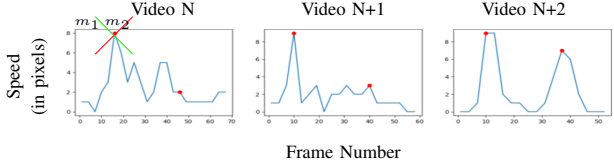
Figure \ref{fig:movement} shows 3 graphs from arbitrary videos ($N\dots N+2$) from the data set with a sliding window of 3 frames. Every video follows a similar pattern: hands accelerate from the resting position into the signing position, then slow down during the signing and then accelerate again into the resting position. Increasing the window size results in smoother graphs, but the trend remains visible.

In order to perform video segmentation using the speed graphs ($S$) in order to separate epentheses and phonemes, we find the extrema in the graphs where there is a change in the slopes ($m_1$ and $m_2$) at consecutive times ($t_1$ and $t_2$) and discard the frames that come before the first maximum and after the last maximum
\[
\sign(\frac{\partial S}{\partial t_1}) \neq \sign(\frac{\partial S}{\partial t_2})
\]

After discarding the frames, all the data for every data set is split 67$\%$ / 16.5$\%$ / 16.5$\%$ for the training/validation/testing respectively. Also, the testing data is verified manually and 78$\%$ of the samples made it into the final test set.

\begin{figure}[H]
\centering
    \begin{tikzpicture}
    \node[anchor=south west,inner sep=0] at (0,0) {\includegraphics[width=0.8\linewidth,trim={0cm 3cm 0cm 0cm},clip]{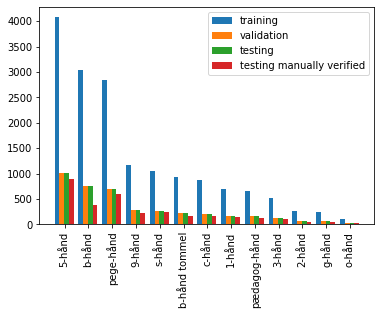}};
    \node[anchor=south west,inner sep=0] at (1,-0.7) {\includegraphics[width=0.65\linewidth,trim={0cm 0cm 0cm 15cm},clip]{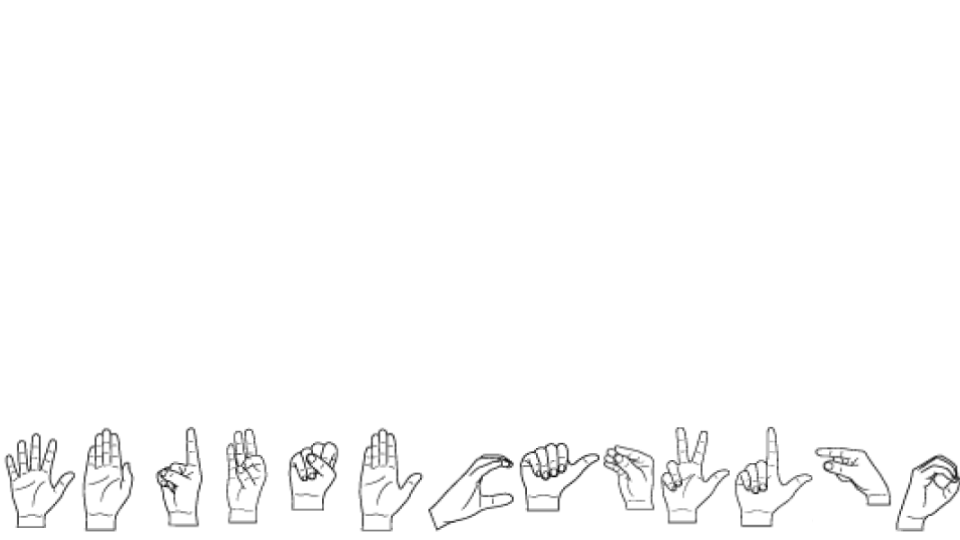}};
    \node[anchor=south west,inner sep=0] at (0,-2.4) {\includegraphics[width=0.8\linewidth,trim={0cm 0cm 0cm 8.2cm},clip]{pics/dataset}};

    \node[very thick] at (-0.2,2) {\rotatebox[origin=c]{90}{\scriptsize Number of frames}};
    
\end{tikzpicture}
\caption{Every generated data set split for model training, validation, and testing. The testing data set was manually verified, resulting in refined testing data set}
\label{fig:splits}
\end{figure}


As it can be seen from the Figure \ref{fig:splits}, after manual verification of the test data set, the $b-hand$ hand shape lost more samples than any other hand shape. This is because the hand shapes and $5-hand$, $b-hand\ tommel$, and $b-hand$ (\includegraphics[trim={4.65cm 8.5cm 24cm 7.4cm},clip,height=\fontcharht\font`\B]{pics/handshapes_2}\includegraphics[trim={0cm 8.5cm 31.5cm 7.4cm},clip,height=\fontcharht\font`\B]{pics/handshapes_2}) are very similar and even the annotation of the original data had many wrong annotations in these three classes.

Since we do not have linguistic background in the sign languages and, in particular, in Danish Sign Language, every image was judged subjectively, but conservatively. In such cases when the hand in the image appeared blurry due to the motion blur or where not all the fingers of the hand were visible due to occlusions, the frames were discarded.

\begin{figure}[H]
\centering
    \begin{tikzpicture}
    \node[anchor=south west,inner sep=0] at (0,0) {\includegraphics[width=0.4\textwidth]{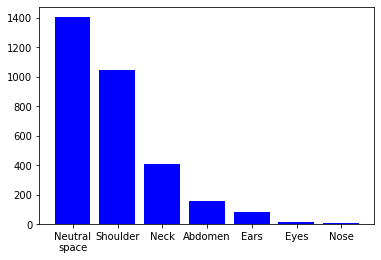}};

    \node[very thick] at (-0.2,2.5) {\rotatebox[origin=c]{90}{\scriptsize Number of frames}};
    
\end{tikzpicture}
    \caption{Number of location instances in the test set}
    \label{fig:tab_stats}
\end{figure}


\begin{figure}[H]
\centering
    \begin{tikzpicture}
    \node[anchor=south west,inner sep=0] at (0,0) {\includegraphics[width=0.4\textwidth]{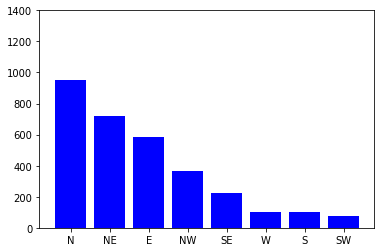}};

    \node[very thick] at (-0.2,2.5) {\rotatebox[origin=c]{90}{\scriptsize Number of frames}};
    
\end{tikzpicture}
    \caption{Number of orientation instances in the test set}
    \label{fig:ori_stats}
\end{figure}

Similarly, location and orientation distributions have been affected after the manual verification of the test data set and can be seen in Figures \ref{fig:tab_stats} and \ref{fig:ori_stats}.

\section{Handshape Classification}\label{sec:dez}
Handshape is defined by the hand configuration, which is made out of the anatomic features, such as carpal, metacarpal, and phalanx bones. OpenPose library provides key-points for these anatomic features as described in Section \ref{sec:related_work}. We used key-points to train Nearest Neighbour, Random Forest, and Feed-Forward Neural Network. We also use raw images and synthesised binary images to train Convolutional Neural Network algorithms to recognise different hand shapes.

K-Fold validation with five folds for the Nearest Neighbour, Random Forest, and Feed-Forward Neural Network methods was used to tune the parameters and report the average prediction accuracies over all the folds. In case when a parameter has alternatives, separated by comma, the parameter in bold show the selected parameters for the best model.

\subsection{Methodology}
For every model used, we report the combination of the parameters that we tried and their results with supporting graphs in Appendix \ref{appendices1}.

\subsubsection{Nearest Neighbour (k-NN)}\label{sec:knn}
is a clustering algorithm, where every new unseen data point is subjected to the $k$-nearest neighbours vote using some distance metric.

\begin{center}
\begin{tabular}{ c c }
 Neighbours & $1,$\textbf{5}$,10,15,20$
\end{tabular}
\end{center}

\subsubsection{Random Forest (RF)}
trains one or more decision trees on sub-samples of the overall data and uses averaging to improve the predictive accuracy and control over-fitting. Decision tree pruning is also used to control over-fitting. Decision tree is a tree-based data structure, where every node learns a decision rule that partitions the overall decision space.

\begin{center}
\begin{tabular}{ c c }
 Maximum Leaf Nodes & $100,300,500,$\textbf{800} \\
 Number of Estimators & $5,10,15,20,25,$\textbf{30} \\
 Maximum Tree Depth & $5,10,15,$\textbf{20} \\
 Minimum Samples per Leaf & $10,$\textbf{50}$,100$ \\
 Minimum Samples per Split & $10,$\textbf{50}$,100$ \\
 Maximum Features & $0.1$
\end{tabular}
\end{center}

\subsubsection{Feed-Forward Neural Network (FFNN)}
is a data structure with every new layer introducing more non-linearity into the decision space. The error function is reduced over the epochs using stochastic gradient descent optimisation method. 

\begin{center}
\begin{tabular}{ l c }
 & \multirow{3}{*}{\shortstack[c]{\textbf{Input-1$\times$Hidden-Output},\\$Input$-$2\times$$Hidden$-$Output$,\\$Input$-$3\times$$Hidden$-$Output$}} \\
 Structure & \\
 & \\
 Activation Function & $ReLu$ \\
 
 Learning Rate initial & $0.01$ \\
 Cosine annealing & $False$ \\
  
 Optimiser & $Adam$ \\
 $\beta_1$ & $0.9$ \\
 $\beta_2$ & $0.999$ \\
  
 Epochs & $200$ \\ 
 Batch Size & $32$ \\
 
 Validation Fraction & $0.1$ \\
 Testing Fraction & $0$ \\
 
 Data Augmentation & $None$
\end{tabular}
\end{center}

\subsubsection{Convolutional Neural Network (CNN)}
is composed of one or many convolution layers. These layers contain a set of kernels (filters). Kernels are optimised during training and each kernel produces a feature map, which acts as a feature extractor for the raw images. In contrast, classical image processing used hand-engineered kernels (e.g. vertical, horizontal, gaussian, laplacian filters) to transform the raw images. However, learned kernels have shown to be more superior to classical hand-crafted kernels.

\begin{center}
\begin{tabular}{ l c }
 CNN Filters & $8,$\textbf{32} \\
 CNN Kernel Size & $3,$\textbf{5} \\
 Dropout Rate & $0.25$ \\
 
 \multirow{2}{*}{\shortstack[c]{Structure}} & \multirow{2}{*}{\shortstack[c]{$Input-1\times CNN-Output$,\\ \textbf{Input$-$3$\times$ CNN$-$Output} }} \\
 & \\
 Activation Function & $ReLu$ \\
 
 Learning Rate initial & $1e-4,$\textbf{1e$-$2} \\
 Cosine Annealing & $True,$ \textbf{False} \\
 
 Optimiser & $Adam$ \\
 $\beta_1$ & $0.9$ \\
 $\beta_2$ & $0.999$ \\
 
 Epochs & $200$ \\
 Batch size & $32$ \\
 
 Validation Fraction & $0.2$ \\
 Testing Fraction & $0.2$ \\
 
 Data Augmentation & $(feature$-$wise)$ $normalisation$
\end{tabular}
\end{center}

\subsubsection{Transfer Learning\label{sec:cnn}}
is a method for pre-training a model on some data so that it can learn data-specific high-level features and then fine-tuning the same pre-trained model on the new data that may share similar features with the data that the model was pre-trained with. In the case of the data set with the binary images, the model has been pre-trained on the MNIST data set by modifying the size of the input to $128\times128$ to match the hand shape data set input size. The idea behind the pre-training on the MNIST data set was to train the model to learn such features as corners and edges, which could also benefit in classification of the hand shapes on the binary data set.

In the case of the Inception network \cite{DBLP:journals/corr/SzegedyVISW15}, the model has been pre-trained on the ImageNet data set.

The idea behind the transfer learning is to provide a useful weight initialisation, which could result in the training to begin near the local or even global minimum in the search space. This, in turn, allows for much less data to be used to reach the minimum or can result in much faster convergence.

\begin{center}
\begin{tabular}{ l c }
 CNN Filters & \multirow{6}{*}{\shortstack[c]{$Simonyan$ $and$ $Zisserman$ \cite{simonyan2014deep}}} \\
 CNN Kernel Size &  \\
 Convolution Layers &  \\
 Dropout Rate &  \\
 
 Structure &  \\
 Activation Function &  \\
 
 Learning Rate initial & $1e-4$ \\
 Cosine Annealing & $False$ \\
 
 Optimiser & $Adam$ \\
 $\beta_1$ & $0.9$ \\
 $\beta_2$ & $0.999$ \\
 
 Epochs & $100$ \\
 Batch size & $32$ \\
 
 Validation Fraction & $0.2$ \\
 Testing Fraction & $0.2$ \\
 
 Data Augmentation & $(feature$-$wise)$ $normalisation$
\end{tabular}
\end{center}

\subsection{Results}
\bgroup
\def\arraystretch{1.5}
\begin{table}[H]
\caption{Combined accuracy results in $\%$ averaged over three runs on the test set of the different methods on both the images data set (Figure \ref{fig:generated_ds} a-b) and the key-points data set (Figure \ref{fig:generated_ds} c-d)}
\resizebox{0.5\textwidth}{!}{%
\begin{tabu}{l l c|c|c|[2pt]c|c|}
& & \multicolumn{5}{c}{Method} \\ \cline{3-7}
& \multicolumn{1}{c|}{Data} & k-NN & \begin{tabular}[c]{@{}c@{}}RF\end{tabular} & FFNN & CNN &  \begin{tabular}[c]{@{}c@{}}Transfer\\ Learning\end{tabular} \\ \cline{2-7}
\multirow{2}{*}{\rotatebox[origin=c]{90}{\shortstack[c]{Images}}} & \multicolumn{1}{|l|}{Cropped Raw Images} & \multicolumn{3}{c|[2pt]}{---} & 76$\pm$0.010 & \textbf{90$\pm$0.008} \\ \cline{2-7}
& \multicolumn{1}{|l|}{Binary Skeleton Hand Images} & \multicolumn{3}{c|[2pt]}{---} & 73$\pm$0.005 & 73$\pm$0.007 \\ \specialrule{.2em}{0em}{0em}
\multirow{2}{*}{\rotatebox[origin=c]{90}{\shortstack[c]{Key-points\;}}} & \multicolumn{1}{|l|}{Raw Features} & 72$\pm$0.0 & 62$\pm$0.004 & 38$\pm$0.008 & \multicolumn{2}{c|}{---} \\ \cline{2-7}
& \multicolumn{1}{|l|}{Distances Between Features} & 76$\pm$0.0 & 68$\pm$0.002 & 65$\pm$0.001 & \multicolumn{2}{c|}{---} \\ \cline{2-7}
\end{tabu}%
}
\vspace{1em}
\label{tbl:results}
\end{table}
\egroup

Table \ref{tbl:results} shows the overall model performance on the test set using the best settings as described in Sections \ref{sec:knn}-\ref{sec:cnn}. The average results are reported with the standard deviation after the three runs.

The most accurate model is the one that is pre-trained on the ImageNet data set and fine-tuned on the cropped raw images of the hands. Surprisingly, the binary data does not perform as well as expected and pre-training the model on the MNIST data set does not improve the performance of the fine-tuned model. Models trained on the distances between the hand features data set perform better than the same models trained on the raw hand features. This is expected as the distances data is more invariant to changes and contains less features.

\subsubsection{Nearest Neighbour (k-NN)\label{sec:knn_res}} number of neighbours affects the classification accuracy of both raw features and the distance features as is shown in Figure \ref{fig:knn}. With only one neighbour, we see that the model overfits as there is a big difference between the training (raw 99\% and distance 99\%) and the testing (raw 72\% and distance 73\%) accuracies.

\subsubsection{Random Forest (RF)} update rules and the structure affect the generalisation of the model. From Figure \ref{fig:rf} we see that the model overfits if we allow the model to make decision nodes using very little number of samples (e.g. 10) with a big difference between the training (raw 99\% and distance 97\%) and the testing (raw 75\% and distance 74\%) accuracies. As a rule of thumb, the more estimators the model has, the better is the performance which comes at the expense of the training time (training raw 88\% and distance 85\% while testing raw 65\% and distance 66\% accuracies with 30 estimators). Tree depth (training raw 88\% and distance 85\% while testing raw 65\% and distance 66\% accuracies with 20 levels deep) and the maximum number of leaf nodes (training raw 88\% and distance 85\% while testing raw 65\% and distance 66\% accuracies with 800 leaf nodes) have the biggest impact on the accuracy of the model, applied both to the raw and distance hand features, but also contributes the most to the overfitting of the model.

Figure \ref{fig:rf_feature_imp} shows the importance of the features from the data, inferred by the model. Interestingly, index and thumb fingers play an important role in distinguishing the hand shapes using either raw key-points or distances features. The fact that the thumb is an important feature in sign languages is supported by both \cite{ojala2009coarticulation,ann1996relation}.

\subsubsection{Feed-Forward Neural Network (FFNN)} parameters impact the performance of the model. Figure \ref{fig:mlp} shows how different number of hidden layers affects the classification of both raw and distance features. Model, trained and tested on distance features performs better than the model trained and tested on the raw features and having a relatively shallow network performs better than having a deeper network (training raw 28\% and distance 58\% while testing raw 27\% and distance 57\% accuracies with one hidden layer with 100 nodes).

\subsubsection{Convolutional Neural Networks (CNN)\label{sec:cnn_res}} parameters affect the accuracy of the model as shown in Figure \ref{fig:cnn}. Using low number of convolutions (e.g. 1) makes the model to overfit as the difference between the training and test set is relatively big (training 98\% and testing 54\%). If the learning rate is discounted, the model is underfitting as it does not reach the same accuracy level as the model that does not discount the learning rate (training 67\% and testing 64\%). All the other parameters have little affect on the accuracy of the model trained on both raw and binary images.


\section{Orientation Classification}\label{sec:ori}
Section \ref{sec:dez} showed that a fine-tuned model on cropped hand images performed the best as compared to other considered approached. This section takes a look at calculation of the orientation phonological parameter. According to the HamNoSys notation system, orientation has two sub-types, namely, extended finger orientation and palm orientation as mentioned in Table \ref{tab:hamnosys}. In this paper we only look at the extended finger orientation.

\subsection{Methodology}
Total of eight orientations have been used for the extended finger orientation as defined in the HamNoSys notation with each orientation having $45^{\circ}$ movement. Despite the HamNoSys defining more orientations (e.g. towards or away from the body), having 2D data makes it difficult to estimate additional orientations. 

\begin{center}
\begin{tabular}{ c c }
 North & North-East \\
 East & South-East \\
 South & South-West \\
 West & North-West
\end{tabular}
\end{center}

The angle has been calculated using the inverse trigonometric function between the radius and middle finger coordinates.
\begin{gather*}
-\pi/2 < \arctan{(q_y - p_y, q_x - p_x)} < \pi/2, \\
\text{where $q$ and $p$ are the $(x,y)$ coordinates} \\
\text{of radius and middle finger metacarpal bones} \\
\text{with every orientation having $\pi/4$ freedom}
\end{gather*}

\subsection{Results}
\begin{figure}[H]
    \centering
    \includegraphics[width=0.5\linewidth,trim={3cm 1cm 3cm 4cm},clip]{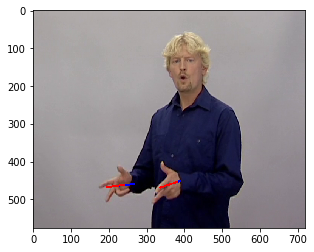}
    \caption{Red line connects detected radius and middle finger metacarpal bones for the calculation of the extended finger orientation}
    \label{sfig:b}
\end{figure}
From Section \ref{sec:related_work}, we can see that none of the previous works modelled orientation phonological parameter. Since our geometry-based approach is dependent on the OpenPose library, the accuracy becomes the same as the accuracy of the library, which is similar to that of the depth sensors \cite{simon2017hand}.

\section{Location Classification}\label{sec:tab}
Sections \ref{sec:dez}-\ref{sec:ori} described how handshape and orientation phonological parameters are modelled. This section takes a look at the location parameter. HamNoSys notation system defines three different location sub-types, namely, hand locations, hand location sides, and hand distances as mentioned in Table \ref{tab:hamnosys}. We focus on hand locations relative to few selected body parts.

\subsection{Methodology}
We have used five locations around the body as opposed to fourty six, defined by the HamNoSys notation system to simplify the detection and to comply with the OpenPose library standards.

\begin{center}
\begin{tabular}{ c c }
 Ears & Eyes \\ 
 Nose & Neck \\
 Shoulder & Abdominal
\end{tabular}
\end{center}

In order to assign the relative hand location, a threshold has to be assigned to how far a centroid of a hand can be from the a specific body location to still be relatively close to that body part. All the distances are measured in pixels and the threshold is set to be $10\%$ of the diagonal of the image frame which is approximately $100$ pixels.

\begin{gather*}
M_r \dots N_r = \lvert q_{m \dots n}- centroid_{right} \rvert \\
M_l \dots N_l = \lvert q_{m \dots n} - centroid_{left} \rvert \\
D = 
 \begin{pmatrix}
  M_r & M_l \\
  \vdots  & \vdots  \\
  N_r & N_l 
 \end{pmatrix}
 \end{gather*}

\noindent Where $q_{m \dots n}$ are the $(x,y)$ position of the body parts, defined by the OpenPose library (e.g. nose, neck, shoulder, elbow, etc.) and the $M_r \dots N_r$ and $M_l \dots N_l$ are the Euclidean distances between the body parts and right and left hand centroids. 

In order to find the body part $B_{right}$ or $B_{left}$ which has the smallest distance to the centroid of the right or left hand, we use 
\[B_{right} = \argmax D_{i,1}\] 
\[B_{left} = \argmax D_{i,2}\]
The distances are then compared to a threshold to determine if a hand is near a particular body part or is in the `neutral signing space' anywhere around the body.

\subsection{Results}
\begin{figure}[H]
    \centering
    \includegraphics[width=0.2\linewidth]{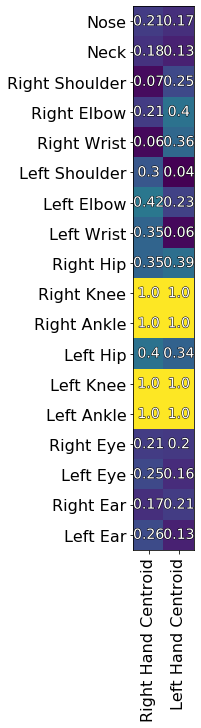}
    \caption{Heatmap of the relative distances between the hand centroids and every body part. From yellow to dark blue colour symbolising long and short distance respectively}
    \label{sfig:heatmap}
\end{figure}

Figure \ref{sfig:heatmap} shows a heatmap for both hand locations relative to all the body parts, normalised by dividing the distances by the diagonal of the frame. Yellow parts, where the distance is $1.0$ means that these body parts are not visible in the frame.

The accuracy of our approach depends on the OpenPose library locating the hand and body parts in an image, which corresponds to the OpenPose library accuracy, which is similar to that of the depth sensors \cite{simon2017hand}.

\section{Co-Dependence of Phonological Parameters}\label{sec:codep}
Sections \ref{sec:dez}-\ref{sec:tab} described how handshape, orientation, and location phonological parameters are modelled. This section is focusing on showing that the two individual phonological parameters are co-dependent. This means that if one phonological parameter is of a certain kind, then it is more likely for another phonological parameter to assume a specific configuration (e.g. it should be more complicated to have a hand above the head and point downwards than pointing downwards while having a hand at the the torso level).

\subsection{Methodology}
\begin{multline*}
\begin{split}
C_{O_N,L_M} & = \begin{pmatrix}
  c_{O_1,L_1} & \hdots & c_{O_1,L_M} \\
  \vdots & \ddots & \vdots  \\
  c_{O_N,L_1} & & c_{O_N,L_M} 
 \end{pmatrix}
\end{split}
\end{multline*}

\begin{multline*}
S_{O_N,L_M} = \sum^{}_{} C_{O_N,L_M} \cap \\(C_{O_i,L_j} \cup \sum^{}_{} O_i \cup \sum^{}_{} L_j)
\end{multline*}

\begin{multline*}
\begin{split}
 C_{2\times 2} = 
 \begin{pmatrix}
  C_{O_i,L_j} & \sum^{}_{} O_i \\
  \sum^{}_{} L_j & S_{O_N,L_M}
 \end{pmatrix}
 \end{split}
\end{multline*}

First, global $C_{O_N,L_M}$ contingency table counts the occurences for both location/orientation variables for every category (e.g. North, North-East, etc. for orientation and Shoulder, Neck, etc. for location) that occur in the collected data. Second, a series of local contingency tables $C_{2\times 2}$ are constructed from the global $C_{O_N,L_M}$ contingency table for every category of every variable as a post-hoc step. Finally, Bonferroni-adjusted $p$-value was used \cite{Bland170} to check if the presence of a particular location/orientation combination ($C_{O_i,L_j}$) in the data set is significant as opposed to other location/orientation combinations ($S_{O_N,L_M}$) by performing the Chi-square test of independence of variables for all the $C_{2\times 2}$ contingency table.

\subsection{Results}
\begin{figure}[H]
\centering
\includegraphics[width=0.5\linewidth,trim={6.5cm 7cm 7.1cm 12.5cm},clip]{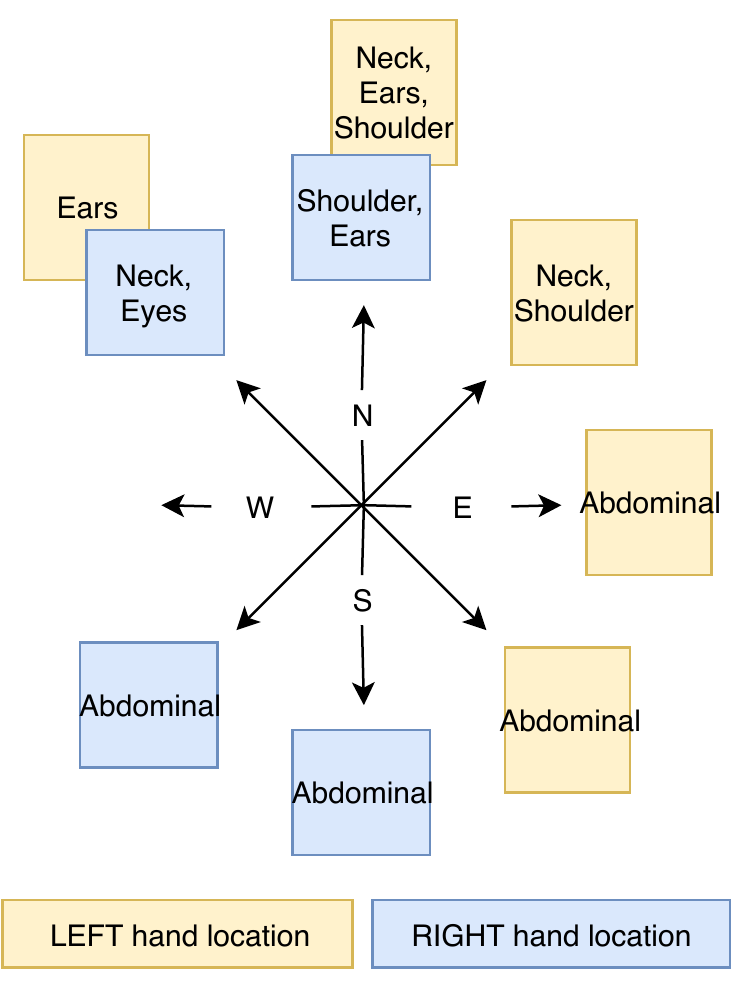}
\caption{Significant \mbox{(Bonferroni-adjusted $p$-value$=0.0029$)} co-dependencies between the location and orientation categories}
\label{fig:result_correlation}
\end{figure}

Figure \ref{fig:result_correlation} shows the significant co-dependence among the orientation and location phonological parameters. The results indicate that it is more common in the data to encounter right hand pointing towards the western side as well as the north and the south, while it is more common for the left hand to point to the eastern side as well as the north and the south, but it is uncommon to point to the eastern side with the left hand and to the western side with the right hand. This has been pointed out by Cooper et al. \cite{cooper2012sign} that only a subset of `comfortable' combinations occurs in practice during signing.

Moreover, for the both hands it is common to point to the northern side at the upper side of the body, while it is common for the both hands to point to the southern side at the lower part of the body.

\section{Multi-Label Fast Region-Based (Fast R-CNN) Convolutional Neural Network}
There has already been an investigation by Awad et al. \cite{5414159} into sharing the features between the individual phonological parameters and how this improves the model. This motivates our choice of an end-to-end model which supports sharing of the learned features among the individual phonological parameters. Moreover, we utilise the knowledge that location and orientation parameters have some co-dependence and explicitly allow one classifier affect another classifier.

\subsection{Methodology}
After the steps taken to classify handshape, orientation, and location phonological parameters in raw images as described in Sections \ref{sec:dez}-\ref{sec:tab} without using the co-dependence information as was described in Section \ref{sec:codep}, an overall annotation file has been generated for training of the single end-to-end multi-label model.

The advantage of the end-to-end model is that it can be trained all at once to describe different phenomena and it is expected that a single model, trained end-to-end should incorporate all the necessary features to describe these different complex phenomena \cite{glasmachers2017limits}.

{\fontsize{7}{4}\selectfont
\begin{lstlisting}[caption={Five samples from the generated dataset},label=generated_sample,captionpos=b]
video_frame x y handedness handshape orientation location
2169_0021.png  224  288      right    1  ne  shoulder
1604_0030.png  192  320      right    1   e   neutral
1249_0017.png  224  256      right    1   n  shoulder
 444_0008.png  160  416      right    1  se   neutral
 182_0040.png   96  288      right    1   n   neutral
\end{lstlisting}
}
Listing \ref{generated_sample} shows five arbitrary instances from the generated annotation file. The first column tells the video and the frame that is being annotated, the second and the third columns tell the $(x,y)$ origin of a bounding box that has shape $128\times 128$ pixels. Fourth to seventh columns tell the handedness, handshape, orientation, and location phonological parameter categories as motivated by the HamNoSys notation system.

\begin{figure}[H]
\centering
\includegraphics[width=0.9\linewidth,trim={3cm 13cm 2cm 4cm},clip]{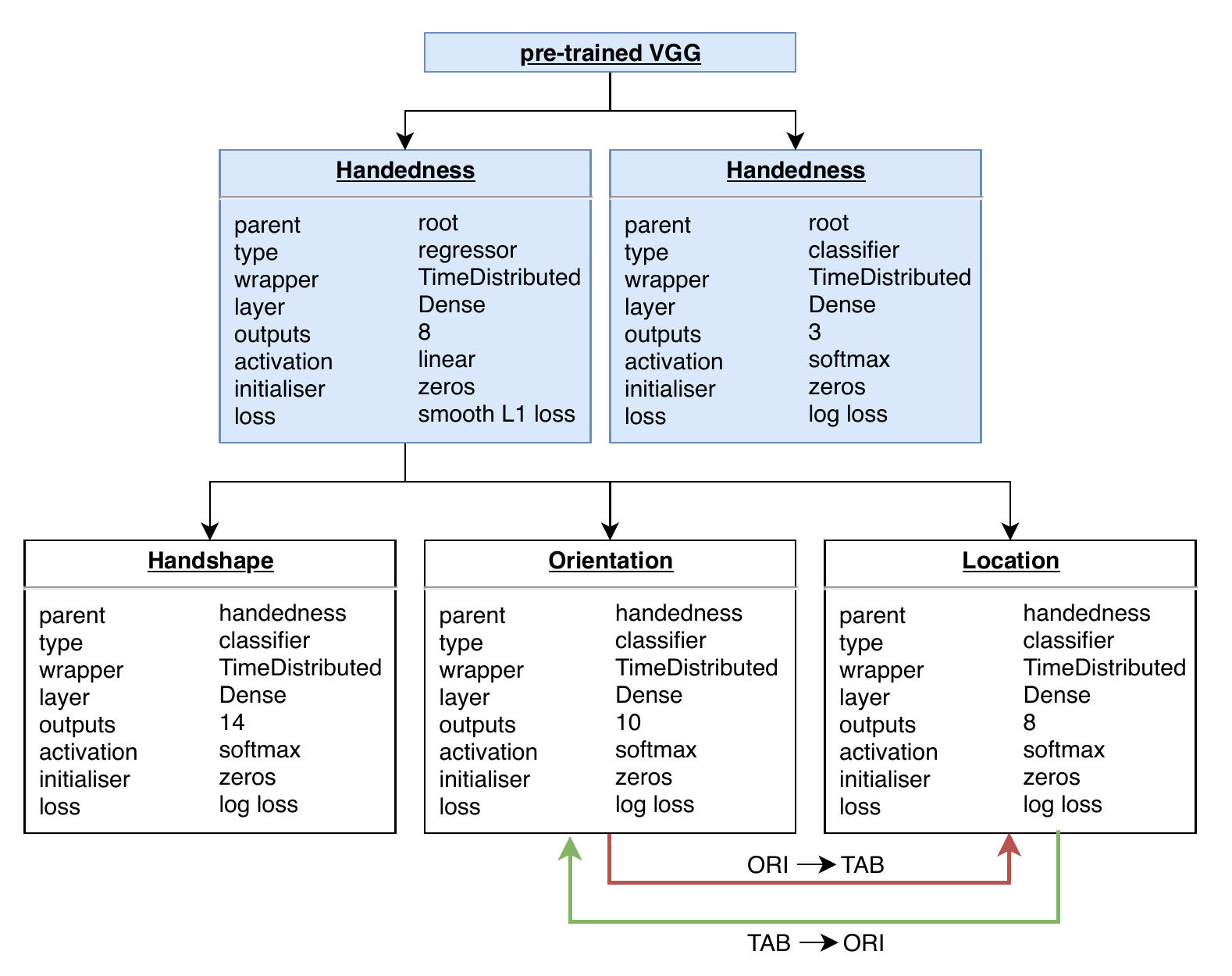}
\caption{Multi-label Fast R-CNN model for detection and classification individual phonological parameters. The model consists of the base model (in blue) that detects and classifies hands. Handshape, orientation, and location correspond to classifiers of individual phonological parameters. Orientation$\rightarrow$Location means that the classification of orientation affects the classification of the location phonological parameter while Location$\rightarrow$Orientation means that the classification of location affects the classification of orientation phonological parameter}
\label{efrcnn}
\end{figure}

Figure \ref{efrcnn} shows the traditional single-label Fast Region-Based Convolutional Neural Network (Fast R-CNN) model (in blue) that has been extended by adding multiple labels into the model, where every label corresponds to a classifier for individual phonological parameter (handshape, orientation, location). The model uses pre-trained network as feature extractor and allows to do both object detection and classification on raw images in a single pass of an input image through the model \cite{girshick2013rich}.

The following parameters have been used to train the network and the validation was performed every 25 epochs to accelerate the training time. Our interest lies in exploiting the label co-dependence that was shown to be present in the data in Section \ref{sec:codep}. 

\begin{center}
\begin{tabular}{ l c }
 CNN Filters &  \multirow{6}{*}{\shortstack[c]{$Girshick$ $et$ $al.$ \cite{girshick2013rich} (in blue)\\$+Handshape,Orientation,Location$\\($Figure$ \ref{efrcnn})}} \\
 CNN Kernel Size &  \\
 Convolution Layers &  \\
 Dropout Rate &  \\
 
 Structure & \\
 Activation Function &  \\
 
 Learning Rate initial & $1e-5$ \\
 Cosine Annealing & $False$ \\
 
 Optimiser & $Adam$ \\
 $\beta_1$ & $0.9$ \\
 $\beta_2$ & $0.999$ \\
 
 Epochs & $100$ \\
 Batch size & $300$ \\
 
 Validation Fraction & $0.2$ \\
 Testing Fraction & $0.2$ \\
 
 Data Augmentation & $normalisation$
\end{tabular}
\end{center}

\subsubsection{Separate\&Independent}\label{sec:first_approach} is our first approach to learning multiple labels that correspond to different classifiers for phonological parameters. All the classifiers are trained independently and sequentially, starting with the handshape classifier. We selected the handshape phonological parameter to be the first one to be learned as it potentially requires very fine features for the correct classification (e.g. phalanges etc) in comparison to other phonological parameters. Once the handshape classifier is trained, all the layers before this classifier are fixed and only the top layers that correspond to classifiers of individual phonological parameters are trained later. This ensures that the features learned by the convolutional layers during the training of the handshape classifier are used by other classifiers as well. Unfortunately, there is no standard quantitative way that we are aware of showing whether the model has learned filters that focus on fine or coarse features of an image.

\paragraph{Separate\&(location$\rightarrow$orientation)} is a variation of the first approach from Section \ref{sec:first_approach} in that it has an additional connection once location classifier is trained to have an effect on the training of the orientation classifier (green arrow location$\rightarrow$orientation in Figure \ref{efrcnn}).

\paragraph{Separate\&(orientation$\rightarrow$location)} is the second variation of the first approach from Section \ref{sec:first_approach} in that it has an additional connection once orientation classifier is trained to have an effect on the training of the location classifier (red arrow orientation$\rightarrow$location in Figure \ref{efrcnn}).

\subsubsection{Joint\&Independent}\label{sec:second_approach} is our second approach where we train all the classifiers at the same time with combined loss function ($Loss_{Handedness}+Loss_{Handshape}+Loss_{Orientation}+Loss_{Location}$). In this case, the learned features have some significance for every single classifier.

\paragraph{Joint\&(location$\rightarrow$orientation)} is the first variation of the second approach from Section \ref{sec:second_approach} in that it has an additional connection once location classifier is trained to have an effect on the training of the orientation classifier (green arrow location$\rightarrow$orientation in Figure \ref{efrcnn}).

\paragraph{Joint\&(orientation$\rightarrow$location)} is the second variation of the second approach from Section \ref{sec:second_approach} in that it has an additional connection once orientation classifier is trained to have an effect on the training of the location classifier (red arrow orientation$\rightarrow$location in Figure \ref{efrcnn}).


\subsection{Results}\label{frcnn:results}

\subsubsection{Separate\&Independent}\label{sec:separately_results}
results in smooth training for every phonological parameter except for the handshape classifier, which starts to overfit after the epoch 50, but then the regularisation is keeping the model from overfitting too much as it can be seen from the Figure \ref{fig:result_trainseparately}. Test set accuracies correspond to 82\%, 88\%, 27\%, and 39\% for the handshape, Handedness, orientation, and location classifiers respectively.

\paragraph{Separate\&(location$\rightarrow$orientation)} results in potentially underfitted classifier as shown in Figure \ref{fig:tab-ori-sep} with accuracy of 13\% on the test set for the orientation classifier.

\paragraph{Separate\&(orientation$\rightarrow$location)} results in relatively a good fit as can be seen on the Figure \ref{fig:ori-tab-sep} with accuracy of 28\% on the test set for the location classifier.

\subsubsection{Joint\&Independent} results in handshape classifier being underfitted as the validation curve strives down as shown in Figure \ref{fig:multi-label-frcnn-training}. The training is slower as opposed to Section \ref{sec:separately_results}. This is understandable since in this approach a combined loss is considered as mentioned in Section \ref{sec:second_approach}. Test set accuracies result in 77\%, 91\%, 32\%, and 56\% for the handshape, Handedness, orientation, and location classifiers respectively.

\paragraph{Joint\&(location$\rightarrow$orientation)} results in potentially underfitted classifier ass can be seen in Figure \ref{fig:tab-ori-sim} with 36\% accuracy on the test set for the orientation classifier.

\paragraph{Joint\&(orientation$\rightarrow$location)} results in potentially underfitted classifier as can be seen in Figure \ref{fig:ori-tab-sim} with 56\% accuracy on the test set for the location classifier.

\subsubsection{Results Comparison}
\begin{center}
\bgroup
\def\arraystretch{1.5}
\begin{table}[H]
\centering
\caption{Test results (in \%) accuracy of the different training variations for the multi-label Fast R-CNN model on the test set}
\begin{adjustbox}{max width=0.48\textwidth}
\begin{tabular}{ |c|c|c|c|c|}
\hline
 \multirow{3}{*}{\shortstack[c]{\textbf{Phonological}\\\textbf{Parameter}\\\textbf{Classifier}}} & \multicolumn{2}{|c|}{\textbf{Separate}} & \multicolumn{2}{|c|}{\textbf{Joint}}  \\ \cline{2-5}
   & \multirow{2}{*}{\shortstack[c]{\textbf{\;\;\;\;\;\;Independent\;\;\;\;\;\;}}} & \multirow{2}{*}{\shortstack[c]{\textbf{location$\rightarrow$orientation}\\\textbf{orientation$\rightarrow$location}}} & \multirow{2}{*}{\shortstack[c]{\textbf{\;\;\;\;\;\;Independent\;\;\;\;\;\;}}} & \multirow{2}{*}{\shortstack[c]{\textbf{location$\rightarrow$orientation}\\\textbf{orientation$\rightarrow$location}}} \\
  & & & & \\ \hline
 Handshape & \multicolumn{2}{c|}{\textbf{82}} & \multicolumn{2}{c|}{77} \\ \hline
 Handedness & \multicolumn{2}{c|}{88} & \multicolumn{2}{c|}{\textbf{91}} \\ \hline
 Orientation & 27 & 13 & 32 & \textbf{36} \\ \hline
 Location & 39 & 28 & \textbf{56} & \textbf{56} \\ \hline
\end{tabular}
\end{adjustbox}
\label{tbl:frcnn-test-results}
\end{table}
\egroup
\end{center}
Table \ref{tbl:frcnn-test-results} shows testing results of the model that was performed on the test set after the model was trained for 100 epochs.

In many cases, different classifiers were underfitted due to the short training time. This can be observed with the handshape classifier when all the classifiers are trained simultaneously (Figure \ref{fig:multi-label-frcnn-training}) and in all the cases when either orientation or location classifiers training are being affected as seen in Figures \ref{fig:tab-ori-sep}, \ref{fig:tab-ori-sim}, and \ref{fig:ori-tab-sim}) except for when the location classifier is affected by the orientation classifier when classifiers are trained separately (Figure \ref{fig:ori-tab-sep}). The difference between the result in Tables \ref{tbl:results} and \ref{tbl:frcnn-test-results} (90\% vs 70\%) can be explained by too short training time in the latter case (100 vs 1000) and the fact that cumulative loss in the latter case would require greater improvements for the handshape classifier for every training batch to improve the overall model.

Overall, the imposed effect on either location or orientation classifiers to facilitate the found co-dependence between the phonological parameters seems to improve the accuracy of these classifiers on the test set as was expected. We can see from the Table \ref{tbl:frcnn-test-results} that when we add a connection between the classifiers that model phonological parameters that are dependent on each other, the performance of these classifiers improves or stays the same, but does not harm the model.

\subsection{Final Multi-Label Model for Individual Phonological Parameters}
Finally, observing that the extra information from a co-dependent classifier could potentially improve the performance of the model as can be seen from Table \ref{tbl:frcnn-test-results}, we have trained the final model where orientation classifier is affected by the location classifier (green arrow location$\rightarrow$orientation in Figure \ref{efrcnn}) and the location classifier is affected by the orientation classifier (red arrow orientation$\rightarrow$location in Figure \ref{efrcnn}) with all the classifiers trained simultaneously. Since many classifiers were underfitted as described in Section \ref{frcnn:results}, the final model was trained for 300 epochs and the optimal results achieved at epoch 200.

\bgroup
\def\arraystretch{1.5}
\begin{table}[H]
\centering
\caption{Related work results compared to our final model results in \% on modelling individual phonological parameters}
\begin{tabular}{ |l|c|c|}
\hline
 \multirow{2}{*}{\shortstack[c]{\textbf{Phonological}\\\textbf{Parameter}}} & \multirow{2}{*}{\shortstack[c]{\textbf{Reported}\\\textbf{result}}} & \multirow{2}{*}{\shortstack[c]{\textbf{Our Final Model}}} \\
 & & \\ \hline
 Handedness & - & 92 \\ \hline
  \multirow{2}{*}{\shortstack[c]{Handshape}} & \multirow{2}{*}{\shortstack[c]{$75$ (Bowden et al. \cite{10.1007/978-3-540-24670-1_30}),\\$63$ (Koller et al. \cite{7780781})}} & \multirow{2}{*}{\shortstack[c]{87}} \\
 & & \\ \hline
 Orientation & - & 68 \\ \hline
 Location & $31$ (Cooper and Bowden \cite{cooper2007large}) & 60 \\ \hline
\end{tabular}
\label{tbl:related_work_final}
\end{table}
\egroup

Table \ref{tbl:related_work_final} shows together the results of the final model compared to the results found in literature. However, handshape results are not directly comparable, since the past work either focused on different sign languages or considered different handshapes.

Figures \ref{fig:handedness-cm}-\ref{fig:tab-cm} shows the confusion matrices for the test set for all the classifiers in the multi-label F-RCNN model. There are very few cases of the misclassification for the handedness classifier with high recall and precision for the both hands. In the case of the handshape classifier, the b-hand tommel (b-hand with a finger to the side) is sometimes misclassified as the b-hand or 5-hand, which are the same hand shapes only without the finger to the side or with all the fingers spread out as can be seen in Figure \ref{fig:video_stats}. Koller et al. \cite{7780781} reports similar results for the handshapes that are similar in both papers. As for the orientation classifier, we see lower recall in the cases of the south-west, south, south-east, and west orientations mostly being confused with the adjacent orientations. Finally, for the location we also see low recall for some cases. For the upper body parts, when a hand is next to the eyes the model most of the time thinks that the hand is next to the ears while if the hand is next to the nose, it is seen as if it was next to the neck. For the lower body parts, when a hand is next to the abdomen the system sees it as if it was in the neutral space.

Figure \ref{fig:frcnn-test-visual} shows three arbitrary frames from the test set, processed by the multi-label F-RCNN model after it was trained for 500 epochs with all the classifiers trained at the same time and extra connections between the location and orientation classifiers to facilitate their co-dependence.

\section{Conclusion}
To conclude, we have shown how different methods handle classification of the handshape and found out that the transfer learning works best on the raw cropped images of the hands. We have discovered that the two phonological parameters (location and orientation) are dependent on each other that reflects the naturalness of the human motion when signing. We tried incorporating this dependency into our overall multi-label model for recognising different phonological parameters and noticed a slight improvement in the performance of the model. 

\subsection{Future Work}
Additional label for movement phonological parameter should be added to the multi-label F-RCNN model to classify hand movement. For this label, memory has to be introduced into the model. Additionally, to improve the classification of the location phonological parameter, the output from the handedness regressor should be an input to the location classifier, which would tell the position of the hands' bounding boxes. Despite the high accuracy of the handshape label, we require fewer labels to comply with the HamNoSys notation. Therefore, the future work will reduce the classification to a smaller set of handshapes and instead classify whether the hand shape has bent fingers and whether the hand shape has extended thumb or not.

In the future, the model will be used to feed the HamNoSys categories, collected from the output to an human-like avatar which will be able to replicate the sign based on the HamNoSys categories.


\ifCLASSOPTIONcompsoc
  \section*{Acknowledgments}
\else
  \section*{Acknowledgment}
\fi
This work was supported by the Heriot-Watt University School of Engineering \& Physical Sciences James Watt Scholarship and Engineering and Physical Sciences Research Council (EPSRC), as part of the CDT in Robotics and Autonomous Systems at Heriot-Watt University and The University of Edinburgh (Grant reference EP/L016834/1)


\clearpage
\newpage
\appendices
\section{}\label{appendices1}
\begin{tabular}{ @{}p{1cm}@{} l l l }
 \hspace{-1.5em}\raisebox{2.5\normalbaselineskip}[0pt][0pt]{\rotatebox[origin=c]{90}{Raw}} & \hspace{-1.5em}\includegraphics[width=0.25\linewidth]{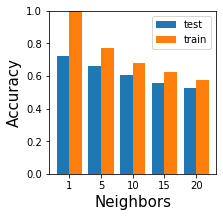} & \raisebox{2.5\normalbaselineskip}[0pt][0pt]{\rotatebox[origin=c]{90}{Distance}} & \includegraphics[width=0.25\linewidth]{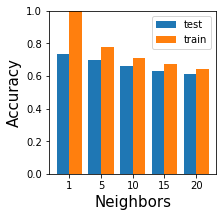} 
\end{tabular}
\begin{figure}[H]
    \caption{k-NN with different k is applied to both raw and distance hand features}
    \label{fig:knn}
\end{figure}

\begin{center}
\begin{tabular}{ @{}p{1cm}@{} @{}p{8cm}@{} }
 \raisebox{3.2\normalbaselineskip}[0pt][0pt]{\rotatebox[origin=c]{90}{Raw}} & \includegraphics[width=\linewidth]{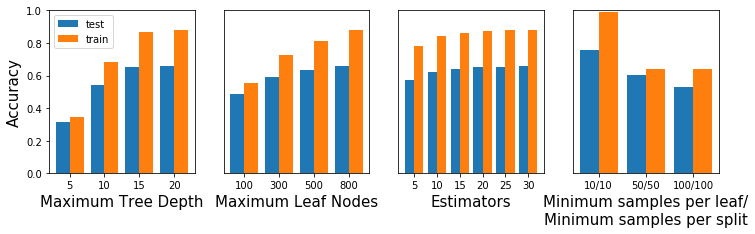} \\ 
 \raisebox{3\normalbaselineskip}[0pt][0pt]{\rotatebox[origin=c]{90}{Distance}} & \includegraphics[width=\linewidth]{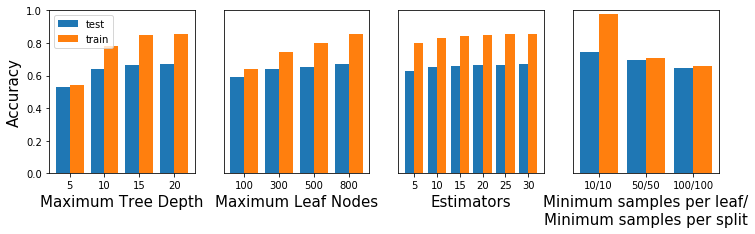}
\end{tabular}
\end{center}
\begin{figure}[H]
    \caption{Random forest with different parameters is applied to both raw and distance hand features}
    \label{fig:rf}
\end{figure}

\begin{tabular}{ @{}p{1cm}@{} l l l }
 \hspace{-1.5em}\raisebox{3.5\normalbaselineskip}[0pt][0pt]{\rotatebox[origin=c]{90}{Raw}} & \hspace{-1.5em}\includegraphics[width=0.25\linewidth]{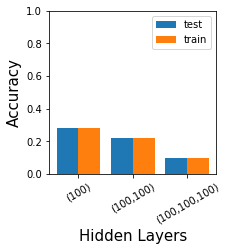} & \raisebox{3.5\normalbaselineskip}[0pt][0pt]{\rotatebox[origin=c]{90}{Distance}} & \includegraphics[width=0.25\linewidth]{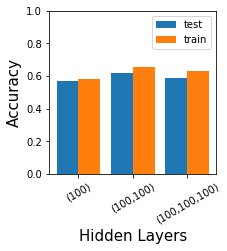}
\end{tabular}
\begin{figure}[H]
    \caption{Feed-forward neural network is applied to both raw and distance hand features}
    \label{fig:mlp}
\end{figure}

\begin{tabular}{ @{}p{1cm}@{} @{}p{8cm}@{} }
 \hspace{-1.5em}\raisebox{2.2\normalbaselineskip}[0pt][0pt]{\rotatebox[origin=c]{90}{Raw}} & \hspace{-1.5em}\includegraphics[width=\linewidth]{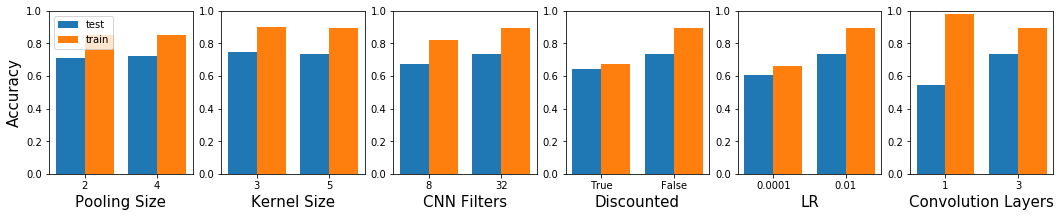}
\end{tabular}
\begin{figure}[H]
    \caption{Convolutional neural networks parameters affect the model accuracy}
    \label{fig:cnn}
\end{figure}

\begin{tabular}{ @{}p{0.5cm}@{} @{}p{4cm}@{} l }
  &  &  \\ 
  
 & &  \\ 
 
 \raisebox{-5\normalbaselineskip}[0pt][0pt]{\rotatebox[origin=c]{90}{Raw}} & \multirow{12}{*}{\includegraphics[width=0.8\linewidth,trim={0cm 0cm 0cm 0.9cm},clip]{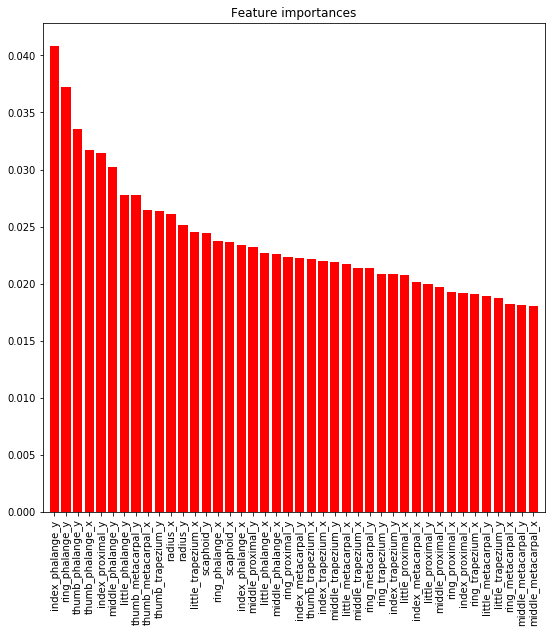}} & \scriptsize \multirow{12}{*}{\shortstack[l]{index\_phalange\_y $(0.042352)$\\ring\_phalange\_y $(0.035244)$\\thumb\_phalange\_y $(0.033687)$\\thumb\_phalange\_x $(0.031453)$\\index\_proximal\_y $(0.031392)$\\middle\_phalange\_y $(0.030922)$\\thumb\_metacarpal\_y $(0.027622)$\\little\_phalange\_y $(0.027398)$\\radius\_x $(0.027240)$\\thumb\_metacarpal\_x $(0.025710)$\\ \; \\ {\normalsize\dots}}}\\ 
 
 & & \\ 
 & & \\ 
  & & \\ 
   & & \\ 
    & & \\ 
     & & \\ 
      & & \\ 
       & & \\ 
        & & \\ 
  &  & \scriptsize \multirow{12}{*}{\shortstack[l]{index\_middle $(0.086312)$\\index\_ring $(0.074342)$\\thumb\_little $(0.072266)$\\thumb\_middle $(0.071572)$\\thumb\_index $(0.071267)$\\ \; \\ {\normalsize\dots}}} \\ 
  
 & &  \\ 
 
  \raisebox{5\normalbaselineskip}[0pt][0pt]{\rotatebox[origin=c]{90}{Distance}} & \includegraphics[width=0.8\linewidth,trim={0cm 0cm 0cm 0.9cm},clip]{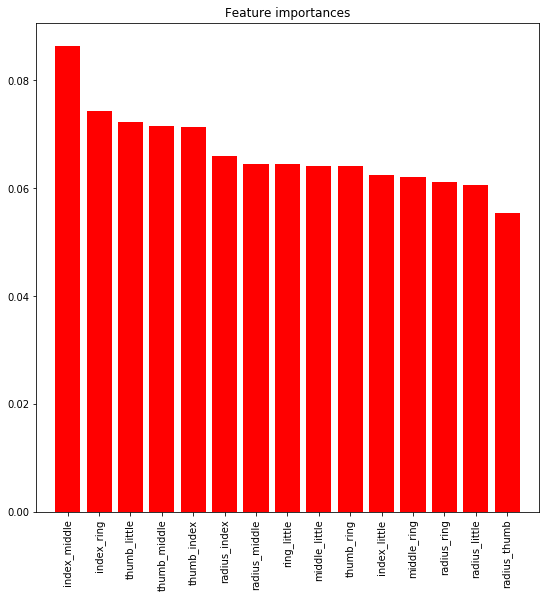} & \\ 
 
 & & \\ 
 & & \\ 
  & & \\ 
   & & \\ 
    & & \\ 
     & & \\ 
      & & \\ 
       & & \\ 
        & & \\ 

\end{tabular}
\vspace{-9em}
\begin{figure}[H]
    \caption{Feature Importances}
    \label{fig:rf_feature_imp}
\end{figure}

\clearpage
\newpage

\section{}\label{appendices15}
\begin{minipage}[H]{0.35\textwidth}
\begin{center}
\includegraphics[width=\linewidth]{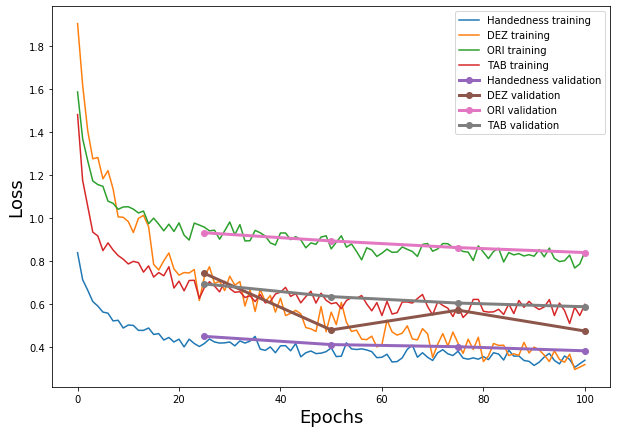}
\label{fig:multi-label-frcnn-training_separately}
\vspace{-2em}
\begin{figure}[H]
    \caption{Training and validation process of the multi-label F-RCNN model for 100 epochs with every label classifier trained separately with shared weights being fixed after the first (Handshape) classifier is trained}
    \label{fig:result_trainseparately}
\end{figure}
\end{center}

\begin{center}
    \includegraphics[width=\linewidth]{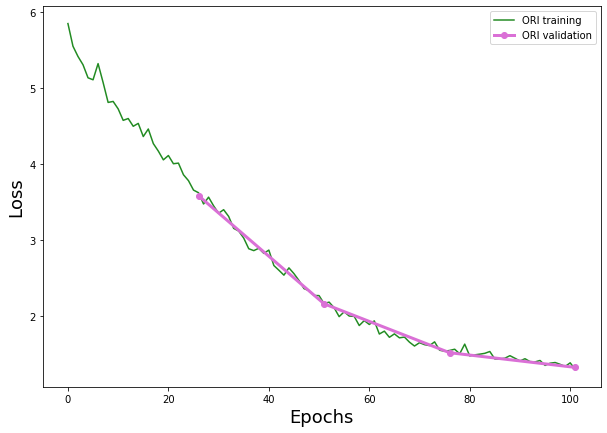}
    \begin{figure}[H]
        \caption{Training and validation process of the orientation classifier affected by the pre-trained location classifier (location$\rightarrow$orientation) with every label classifier trained separately}
        \label{fig:tab-ori-sep}
    \end{figure}
\end{center}

\begin{center}
    \includegraphics[width=\linewidth]{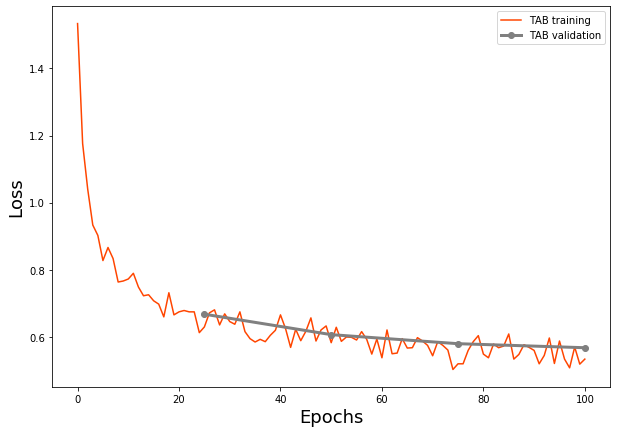}
    \begin{figure}[H]
        \caption{Training and validation process of the location classifier affected by the pre-trained orientation classifier (orientation$\rightarrow$location) with every label classifier trained separately}
        \label{fig:ori-tab-sep}
    \end{figure}
\end{center}
\end{minipage}

\begin{minipage}[H]{0.35\textwidth}
\vspace{7em}
\begin{center}
\includegraphics[width=\linewidth]{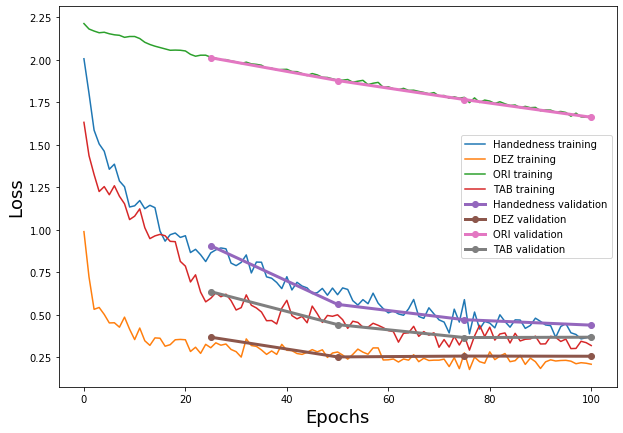}
\vspace{-2em}
\begin{figure}[H]
\caption{Training and validation process of the multi-label F-RCNN model for 100 epochs with all the labels being trained simultaneously}
\label{fig:multi-label-frcnn-training}
\end{figure}
\end{center}

\begin{center}
\vspace{1.2em}
    \includegraphics[width=\linewidth]{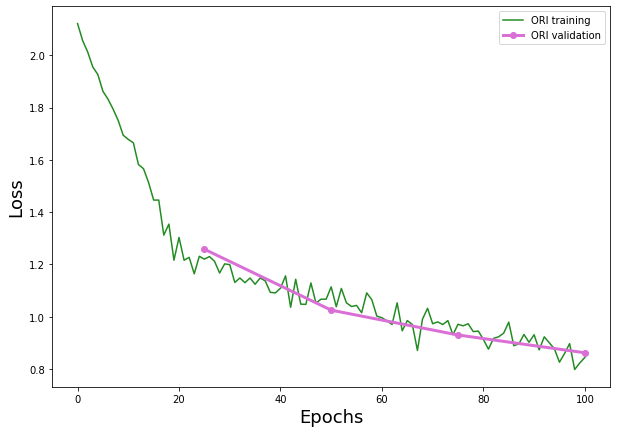}
    \begin{figure}[H]
        \caption{Training and validation process of the orientation classifier affected by the pre-trained location classifier (location$\rightarrow$orientation) with every label classifier trained simultaneously}
        \label{fig:tab-ori-sim}
    \end{figure}
\end{center}

\begin{center}
\vspace{0.5em}
    \includegraphics[width=\linewidth]{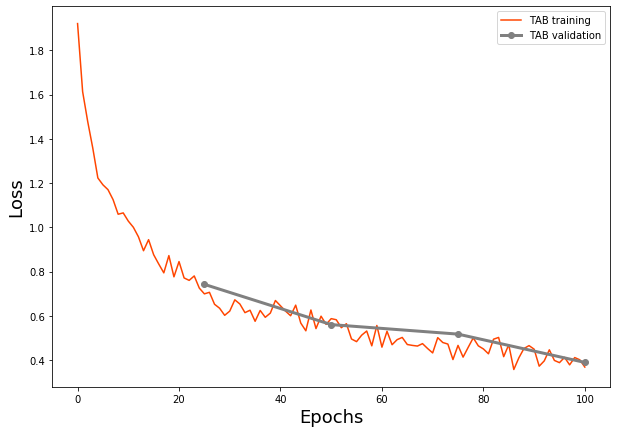}
    \begin{figure}[H]
        \caption{Training and validation process of the location classifier affected by the pre-trained orientation classifier (orientation$\rightarrow$location) with every label classifier trained simultaneously}
        \label{fig:ori-tab-sim}
    \end{figure}
\end{center}

\end{minipage}

\clearpage
\newpage

\section{}\label{appendices17}
\begin{figure}[H]
\centering
\includegraphics[width=0.3\linewidth]{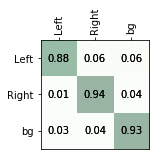}
\caption{Handedness Confusion Matrix on the test set using the final multi-label FRCNN model}
\label{fig:handedness-cm}
\end{figure}

\begin{figure}[H]
\centering
\includegraphics[width=0.8\linewidth]{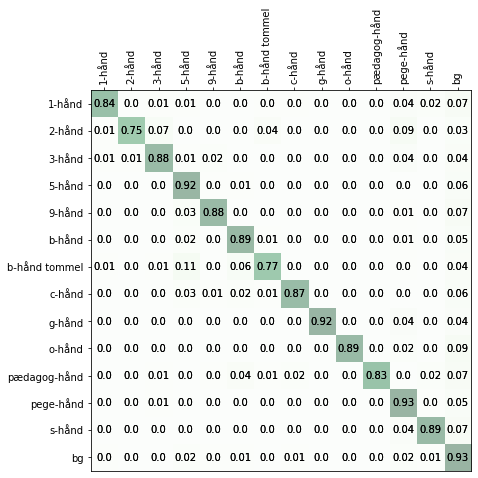}
\caption{Handshape Confusion Matrix on the test set using the final multi-label FRCNN model}
\label{fig:dez-cm}
\end{figure}

\begin{figure}[H]
\centering
\includegraphics[width=0.8\linewidth]{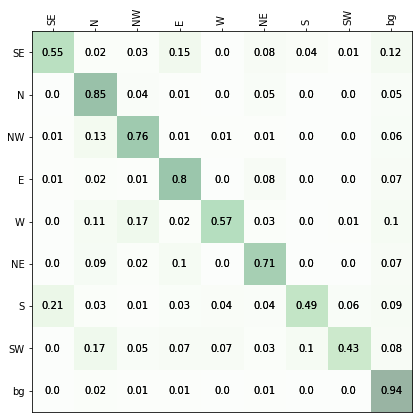}
\caption{Orientation Confusion Matrix on the test set using the final multi-label FRCNN model}
\label{fig:ori-cm}
\end{figure}

\begin{figure}[H]
\centering
\includegraphics[width=0.8\linewidth]{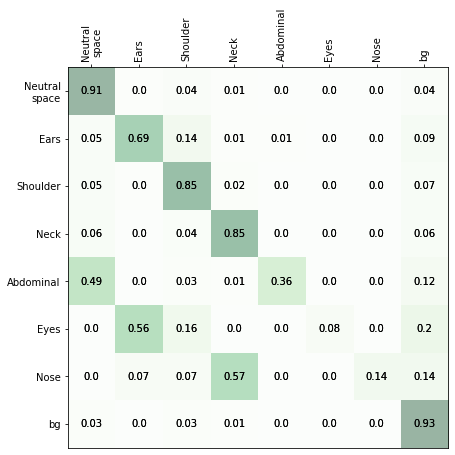}
\caption{Location Confusion Matrix on the test set using the final multi-label FRCNN model}
\label{fig:tab-cm}
\end{figure}

\clearpage
\newpage

\section{}\label{appendices2}
\begin{figure}[H]
    \centering
    \includegraphics[width=0.8\linewidth,trim={4cm 4cm 4cm 2cm},clip]{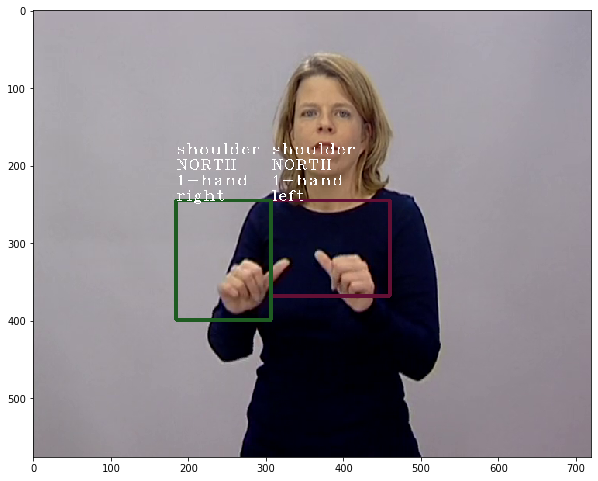}
    \includegraphics[width=0.8\linewidth,trim={5cm 4cm 4cm 2cm},clip]{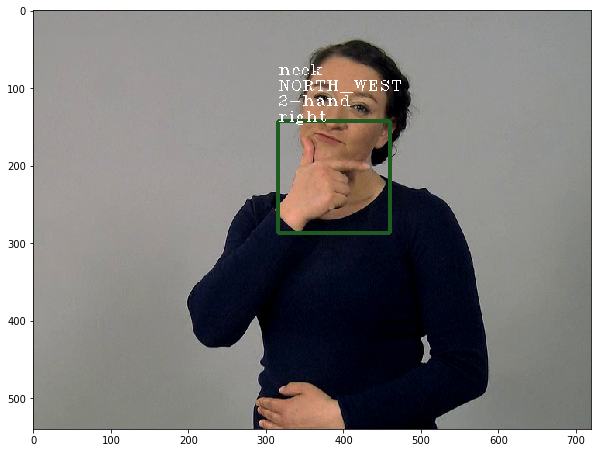}
    \includegraphics[width=0.8\linewidth,trim={4cm 4cm 4cm 2cm},clip]{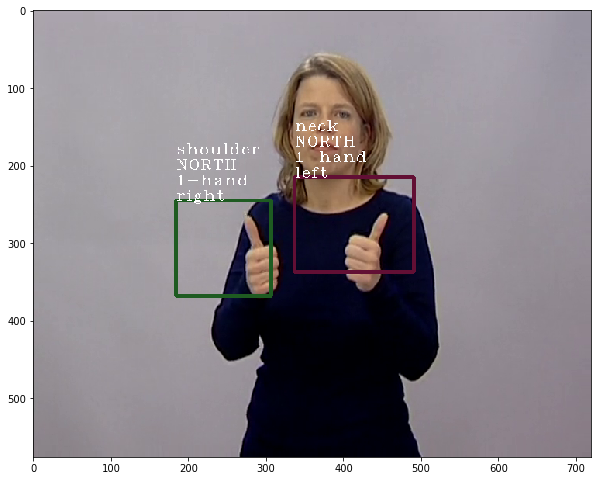}
    \caption{Visualisation of the classification of three arbitrary frames from the test set using the final multi-label FRCNN model}
    \label{fig:frcnn-test-visual}
\end{figure}



%

\clearpage
\newpage
\nocite{*}
\bibliographystyle{amsplain}
\bibliography{bibl.bib}


%

\end{document}